%% file: main.tex
\documentclass[pdflatex,sn-mathphys-num]{sn-jnl}

\usepackage{graphicx}
\usepackage{multirow}
\usepackage{amsmath,amssymb,amsfonts}
\usepackage{xcolor}
\usepackage{booktabs}
\usepackage{array}
\usepackage{flafter}
\usepackage{placeins}
\usepackage{capt-of}
\usepackage{float}
\usepackage{algorithm}
\usepackage{algorithmicx}
\usepackage{algpseudocode}

\newcommand{\mha}{\,\mathrm{mHa}}

\begin{document}

\title[Efficient quantum architecture search through multi-step imagined feedback]{Efficient Quantum Architecture Search through Multi-step Imagined Feedback}

\author[1]{\fnm{Jiayang} \sur{Niu}}

\author[1]{\fnm{Yan} \sur{Wang}}

\author[1]{\fnm{Jie} \sur{Li}}

\author[1]{\fnm{Ke} \sur{Deng}}

\author[1]{\fnm{Azadeh} \sur{Alavi}}

\author[2]{\fnm{Muhammad} \sur{Usman}}

\author*[1]{\fnm{Yongli} \sur{Ren}}\email{yongli.ren@rmit.edu.au}

\affil[1]{\orgdiv{School of Computing Technologies}, \orgname{RMIT University}, \orgaddress{\city{Melbourne}, \postcode{3000}, \state{VIC}, \country{Australia}}}

\affil[2]{\orgdiv{Faculty of Information Technology}, \orgname{Monash University}, \orgaddress{\city{Clayton}, \postcode{3800}, \state{VIC}, \country{Australia}}}

\abstract{
Quantum circuit architecture strongly influences the accuracy of variational quantum algorithms. Reinforcement-learning-based quantum architecture search (RL-QAS) enables sequential circuit construction, but requires repeated variational quantum eigensolver (VQE) optimizations to evaluate them. Adding a legal
gate determines the resulting circuit exactly, but evaluating its energy remains
costly.
Here, we introduce DreamQAS, a model-based reinforcement learning framework that reuses VQE feedback for policy learning. 
DreamQAS combines exact circuit transitions with an ensemble of feedback predictors to generate multi-step imagined trajectories, enabling learning without VQE optimization at every step. 
Ensemble disagreement and selective verification constrain unreliable predictions. 
Across five molecular benchmarks, DreamQAS reaches matched energy-error targets with 1.5--3.4 times fewer VQE optimizations than a non-imagining counterpart and achieves lower mean errors under matched training-episode budgets.
Controlled comparisons support exact transitions and multi-step learning, while reliability analyses identify inaccurate predictions. These findings demonstrate efficient feedback reuse in RL-QAS.
}


\maketitle

\input{section/introduction}
\input{section/results}
\input{section/discussion}

\input{section/methods}
\backmatter

\bmhead{Data availability} All data supporting the findings of this study were generated computationally. Molecular geometries, basis sets, active spaces and Hamiltonian specifications are provided in the Supplementary Information. Molecular-generation files, resolved run configurations and experimental outputs are available in the public repository at \url{https://github.com/jiayangniu/DreamQAS.git}.

\bmhead{Code availability} The implementation of DreamQAS, baseline configurations and scripts used to reproduce the experiments, tables and figures are available at \url{https://github.com/jiayangniu/DreamQAS.git}.

\bmhead{Acknowledgements} This research was supported by the ARC Discovery Project (DP210100743) and the RMIT Advanced Computing Ecosystem (RACE). We thank RACE for providing the high-performance computing resources and computational infrastructure used in this work.

\bmhead{Author contributions} J.N. conceived the study, developed the method and software, conducted the experiments, analyzed the results and wrote the manuscript. Y.W. contributed to the reinforcement-learning methodology and study design. J.L., K.D., A.A., M.U. and Y.R. supervised the work and reviewed and edited the manuscript. All authors approved the final manuscript.

\bmhead{Competing interests} The authors declare no competing interests.

\bibliography{sn-bibliography}

\input{section/supplementary}
\end{document}

%% file: section/introduction.tex
\section{Introduction}\label{sec1}

Finding the lowest-energy state of a molecule, its ground state, is a central problem in quantum chemistry and one of the most promising applications of near-term quantum computers~\cite{cerezo2021variational,tilly2022variational}. The variational quantum eigensolver (VQE) approaches this problem by preparing a trial state with a parameterized quantum circuit and adjusting the parameters to lower its energy~\cite{peruzzo2014variational}. How well this works depends on the circuit's architecture---which gates it contains and in what order---because the architecture determines which states can be reached and how easily the parameters can be optimized~\cite{kandala2017hardware,du2022quantum}. No single architecture suits every molecule, so quantum architecture search (QAS) aims to design circuits automatically~\cite{martyniuk2024quantum,yu2025knowledge}. 
{
\color{black}Some QAS methods build circuits step by step, iteratively selecting operators by a fixed criterion such as the energy gradient~\cite{grimsley2019adaptive,tang2021qubit,feniou2023overlap}.
Others work with whole circuits: they make the choice of gates continuous so that it can be optimized with gradients~\cite{zhang2022differentiable,wu2023quantumdarts}, rank candidate circuits with inexpensive estimates instead of optimizing each one~\cite{he2024training,anagolum2024elivagar}, or train a model to generate complete circuits~\cite{nakaji2024generative,furrutter2024quantum}.
}

In reinforcement-learning QAS (RL-QAS)~\cite{kundu2026reinforcement,ikhtiarudin2025benchrl}, circuits are also built step by step, but the rule for choosing each gate is learned from experience rather than fixed in advance~\cite{ostaszewski2021reinforcement,wang2024rnn,zhu2023quantum,patel2024curriculum,niu2025hybrid}. This learned rule, called a policy, improves by trial and error: it adds one gate at a time, and the circuits it builds are evaluated to show whether its choices were good. In the RL-QAS methods considered here, this evaluation uses the circuit's energy: after every gate addition, the circuit's parameters are optimized with VQE, and the energy reached shows whether the gate helped. Because learning takes many trials, these VQE optimizations become a major cost of the search~\cite{lu2023qas,niu2026hamqasbench}. Yet which circuit a sequence of gate additions produces is always known exactly; only how good that circuit is requires VQE. Over a search, these costly optimizations also leave a record of how each circuit was built, gate by gate, and the energy reached at every step. Could a model learn from this record to predict how good new circuits are, so that the policy can practice on these predictions instead of running VQE at every step?

% Learned models have previously been used to reduce the cost of searching for quantum circuits, primarily by predicting circuit quality to rank or select candidates~\cite{zhang2021neural,he2023gnn,deng2023progressive,soloviev2024trainability,martyniuk2025benchmarking}. A related approach uses a model to evaluate possible future decisions before selecting an action, as in model-based planning methods such as MuZero~\cite{schrittwieser2020mastering}, which has also been applied to quantum encoding-circuit design~\cite{rapp2025reinforcement}. A third approach, developed in model-based reinforcement learning, trains policies on imagined trajectories generated by a learned world model~\cite{ha2018world,hafnerdream,hafner2025mastering}. Such models typically approximate both state transitions and feedback. For sequential circuit construction, however, learning the transition is unnecessary: the circuit produced by each legal gate addition can be determined exactly. This suggests a more focused use of imagination in which only the expensive evaluation feedback is predicted, and those predictions are used to train the policy rather than solely to select circuits.

{\color{black}
Learned models have been used in two main ways to reduce the cost of quantum architecture search. Predictor-based QAS methods predict the quality of candidate circuits and use these predictions to decide which circuits to evaluate or keep~\cite{zhang2021neural,he2023gnn,deng2023progressive,soloviev2024trainability,martyniuk2025benchmarking}. Planning-based methods instead use a learned model to evaluate possible next decisions before choosing each gate, as in MuZero~\cite{schrittwieser2020mastering}, which has been applied to the design of encoding circuits for quantum machine learning~\cite{rapp2025reinforcement}. In both cases, predictions mainly guide which circuits or gates are chosen, rather than generating additional circuit-building experience on which a policy can be trained. Generating such experience is the idea behind world models in model-based reinforcement learning, where a policy is trained on imagined trajectories produced by a learned model of both state transitions and feedback~\cite{ha2018world,hafnerdream,hafner2025mastering}. For circuit construction, this idea can be applied in a more focused way: because the circuit produced by each legal gate addition is known exactly, only the expensive evaluation feedback needs to be learned, and imagined circuits can be built with the exact gate rule.
}

\begin{figure}[!ht]
  \centering
  \includegraphics[width=\linewidth]{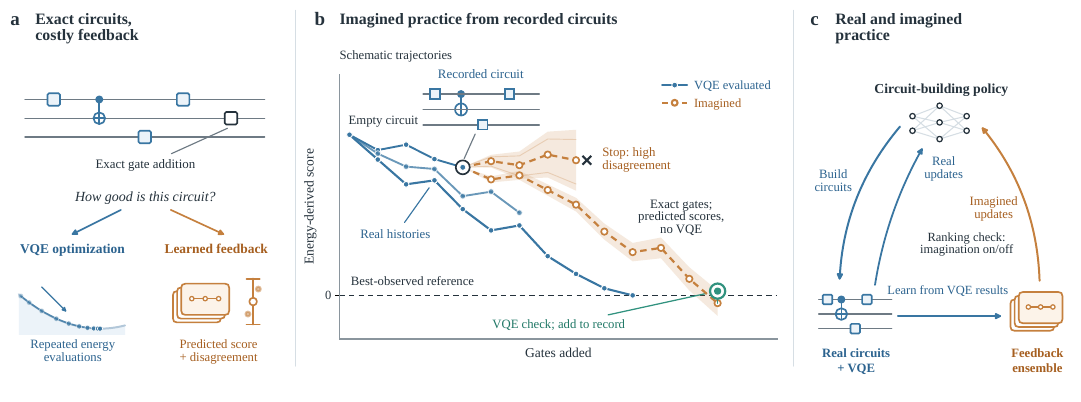}
  % \caption{\textbf{DreamQAS learns from past VQE evaluations through imagined practice.} (\textbf{a}) Adding a gate changes the circuit exactly, whereas its quality requires a costly VQE optimization; DreamQAS replaces this feedback with the predicted score and disagreement of the feedback ensemble. (\textbf{b}) Recorded VQE histories (blue) are extended by imagined sequences (orange) that use exact gate additions and predicted scores without VQE; shading indicates ensemble disagreement. Scores are measured against the best energy observed so far, the frontier $F$ (Eq.~\eqref{eq:res_score}). A sequence stops when disagreement is high, and selective verification checks chosen imagined circuits with VQE and adds them to the record (green). (\textbf{c}) VQE results update the feedback ensemble, while real and imagined practice train the circuit-building policy; imagined updates are used only while the ensemble passes the ranking check.}
  {\color{black}
  \caption{\textbf{DreamQAS learns from past VQE evaluations via imagined practice.} \textbf{(a)}: Adding a gate changes the circuit exactly, but its quality requires a costly VQE optimization; DreamQAS instead predicts this feedback, as a score and a disagreement, with a feedback ensemble. \textbf{(b)}: Recorded VQE histories (blue) are extended by imagined sequences (orange): exact gate additions scored by the ensemble without VQE. Scores are relative to the best energy observed so far (dashed line). A sequence stops when disagreement (shading) is high, and selected imagined circuits are checked by VQE and added to the record (green). \textbf{(c)}: VQE results train the ensemble, while real and imagined practice train the circuit-building policy; a ranking check switches imagination on or off.}
  \label{fig:dreamqas_overview} }
\end{figure}

Here we introduce DreamQAS, a reinforcement-learning framework that combines exact circuit construction with learned VQE feedback~(Fig.~\ref{fig:dreamqas_overview}).
A recurrent ensemble is trained on previously evaluated gate sequences to predict an energy-derived score for new circuit architectures. The policy alternates between real episodes, in which partial circuits are evaluated by VQE, and imagined episodes, in which recorded partial circuits are extended through exact gate additions and scored by the ensemble without additional VQE optimization~\cite{sutton1991dyna,janner2019trust}. Multi-step imagined trajectories provide policy-learning signals from predicted improvements across successive gate choices. To limit the influence of inaccurate feedback, DreamQAS enables imagined updates only when the ensemble passes a ranking-based reliability check, penalizes or terminates trajectories with high ensemble disagreement, and selectively verifies promising or uncertain imagined circuits using VQE~\cite{lakshminarayanan2017simple,osband2018randomized,chua2018pets,yu2020mopo,kidambi2020morel}. The ensemble learns scores relative to the best energy observed during search, so its learning targets do not require the exact molecular ground-state energy.

We evaluated DreamQAS on five ground-state benchmarks for LiH and BeH$_2$ with 4 to 10 qubits, comparing it with other RL-QAS methods and with QAS methods from other families. Compared with the same method without imagination, practicing in imagination reduced the VQE optimizations needed to reach the lowest error level attained by both by a factor of 1.5--3.4 (Fig.~\ref{fig:vqe_efficiency}d). DreamQAS also had the lowest mean energy error of all compared methods on all five benchmarks (Table~\ref{tab:main_quality}). Controlled tests on two LiH benchmarks showed that each of the design choices described above outperformed its alternative: exact rather than learned circuit updates, policy training rather than direct circuit selection, and multi-step rather than one-step imagination (Fig.~\ref{fig:design_controls}). The ensemble's ranking of candidate next gates became more consistent with that of VQE during training, though it remained far from exact (Fig.~\ref{fig:action_ranking}), and its disagreement identified unreliable predictions on the three benchmarks tested (Fig.~\ref{fig:risk_coverage}). 
Together, these findings demonstrate that costly VQE feedback can be reused as additional sequential learning experience, while highlighting the importance of how predicted feedback is used and controlled.

%% file: section/results.tex
\section{Results}\label{sec:results}

\subsection{DreamQAS reuses VQE feedback through imagined circuit construction}\label{sec:dreamqas_overview}

DreamQAS addresses the cost of repeated variational quantum eigensolver (VQE) optimizations required to evaluate intermediate architectures during reinforcement-learning-based quantum architecture search (RL-QAS). 
Its central idea is to separate two components of sequential circuit design: constructing a new architecture and evaluating its quality. 
When a gate is added, the resulting architecture is determined exactly by the current gate sequence and selected action. 
In contrast, determining the achievable energy of that architecture generally requires optimizing its circuit parameters using VQE. 
DreamQAS preserves the exact construction rule and learns to predict the expensive evaluation feedback from previously optimized circuits; together, these form its world model, in which only the feedback is learned (Fig.~\ref{fig:dreamqas_overview}).
% DreamQAS preserves the exact construction rule and learns to predict the expensive evaluation feedback from previously optimized circuits (Fig.~\ref{fig:dreamqas_overview}).

% The framework alternates between real and imagined circuit-building experience. During real episodes, a policy selects gates sequentially and VQE evaluates the resulting partial circuits after parameter optimization. 
% Observed energies provide policy-learning rewards and are stored with the corresponding gate sequences. An ensemble of recurrent predictors learns to estimate the quality of circuits without additional VQE evaluations (Fig.~\ref{fig:dreamqas_overview}a). 
% It predicts energy-derived scores relative to the best energy observed during search, rather than using the exact molecular ground-state energy as a learning target. The ensemble's averaged recurrent representation also serves as the input to the circuit-building policy.

The framework alternates between real and imagined circuit-building experience. During real episodes, a policy selects gates sequentially and VQE evaluates the resulting partial circuits after parameter optimization. Observed energies provide policy-learning rewards and are stored with the corresponding gate sequences {\color{black}in a training record. This record is used to train an ensemble of recurrent predictors, the feedback ensemble, to estimate the quality of new circuits} without additional VQE evaluations (Fig.~\ref{fig:dreamqas_overview}a). It predicts energy-derived scores relative to the best energy observed during search {\color{black} (the frontier)}, rather than using the exact molecular ground-state energy as a learning target. The ensemble's averaged recurrent representation also serves as the input to the circuit-building policy.

During imagined episodes, the policy begins from a previously evaluated partial circuit and extends it through successive gate additions using the exact construction rule (Fig.~\ref{fig:dreamqas_overview}b). 
Instead of optimizing each imagined circuit, the ensemble predicts its feedback. 
DreamQAS generates trajectories of up to $H=15$ imagined gate additions and uses predicted improvements across these trajectories to update the policy. 
This allows an improvement predicted after several gate additions to influence earlier decisions. 
Unlike methods that primarily use a predictor to select promising circuits, DreamQAS uses predicted feedback to generate additional sequential policy-learning experience.

% The usefulness of imagined learning depends on prediction reliability. DreamQAS therefore employs three controls (Fig.~\ref{fig:dreamqas_overview}b,c): 
% a ranking-based check enables imagined updates only when the ensemble sufficiently agrees with VQE on recently evaluated circuit orderings; ensemble disagreement penalizes uncertain feedback and can terminate imagined trajectories; 
% and selective verification periodically evaluates promising or uncertain imagined circuits using real VQE optimization. 
% Verified results are added to the training record, and all verification optimizations are included in the reported VQE costs. 
% The score transformation, imagined rewards, policy objective and reliability thresholds are specified in Methods and Supplementary Table~\ref{tab:supp_hyperparameters}.

The usefulness of imagined learning depends on prediction reliability. DreamQAS therefore employs three {\color{black}reliability} controls (Fig.~\ref{fig:dreamqas_overview}b,c): 
a ranking-based check enables imagined updates only {\color{black}after a warm-up period of real episodes and while} the ensemble sufficiently agrees with VQE on recently evaluated circuit orderings; ensemble disagreement penalizes uncertain feedback and can terminate imagined trajectories; 
and selective verification periodically evaluates promising or uncertain imagined circuits using real VQE optimization. 
Verified results are added to the training record, and all verification optimizations are included in the reported VQE costs. 
The score transformation, imagined rewards, policy objective and reliability thresholds are specified in Methods and Supplementary Table~\ref{tab:supp_hyperparameters}.

Two internal controls help distinguish the contributions of the ensemble and imagined learning. NoImag retains the ensemble and policy representation but removes imagined policy updates and selective verification. 
DreamQAS-RL additionally removes the ensemble and trains directly from real VQE feedback. 
Thus, DreamQAS versus NoImag measures the combined imagined-learning and verification process, while NoImag versus DreamQAS-RL examines the benefit of the ensemble representation without imagination.

\subsection{Imagined feedback reduces VQE evaluation cost and improves circuit quality}\label{sec:quality_efficiency}

We test DreamQAS on five molecular ground-state tasks---LiH-4q, BeH$_2$-6q, LiH-6q, BeH$_2$-8q and BeH$_2$-10q---where the suffix gives the number of qubits (Supplementary Table~\ref{tab:supp_search_spaces}). Circuit quality is measured by the energy error, the gap between a circuit's VQE-optimized energy and the exact ground-state energy, in {\color{black}millihartree (mHa)}, and chemical accuracy corresponds to an error of $1.6\mha$; the exact energy is used to report this error, but not in DreamQAS's learning targets or rewards. Search cost is measured in VQE optimizations: each complete optimization of a circuit's parameters counts as one, whereas the energy evaluations within it are not counted separately. In separate timed runs on LiH-4q and LiH-6q, VQE optimizations took $89$--$93\%$ of the training time of DreamQAS, whereas training the ensemble and generating imagined sequences together took $3$--$5\%$ (Supplementary Fig.~\ref{fig:supp_timing}). 

{\color{black}We compare DreamQAS with two internal controls and five published QAS methods spanning different learning and search paradigms. The RL-QAS group comprises the internal controls introduced above, NoImag and DreamQAS-RL,} and two published reinforcement-learning-based methods, CRLQAS~\cite{patel2024curriculum} and HyRLQAS~\cite{niu2025hybrid}. All four RL-QAS baselines are evaluated under the same training-episode budget, with the published methods retaining their original reward formulations. To broaden the comparison beyond reinforcement learning, we further include three methods representing distinct QAS paradigms: the generative GQE~\cite{nakaji2024generative}, the training-free TFQAS~\cite{he2024training}, and the differentiable QuantumDARTS~\cite{wu2023quantumdarts}. These methods follow their respective search procedures and computational budgets {\color{black}(Supplementary Table~\ref{tab:supp_methods})} and are therefore included as reference points for final circuit quality rather than directly matched search-efficiency comparisons. Finally, every method is run with five independent seeds.

\subsubsection{VQE optimizations required to reach matched energy targets.}

Figure~\ref{fig:vqe_efficiency}a--c follows training on three representative tasks; the other two are shown in Supplementary Fig.~\ref{fig:supp_speed}. The horizontal axis counts VQE optimizations as they accumulate, and the vertical axis shows the energy error of the best circuit found in each training episode, averaged over the most recent 100 episodes; curves give the median and interquartile range across seeds. At a fixed error level, the horizontal gap between two curves therefore shows how many more VQE optimizations one method needs to reach that error. To summarize this gap for all five tasks, we choose for each task the tightest error target, from a fixed set, that both DreamQAS and NoImag reach: $0.5$, $0.5$, $12$, $2.5$ and $3\mha$ for LiH-4q, BeH$_2$-6q, LiH-6q, BeH$_2$-8q and BeH$_2$-10q, respectively. A method reaches a target when its median curve stays at or below it for three consecutive reported points, and we record the cumulative number of VQE optimizations at the first of these points.

\begin{figure}[!ht]
  \centering
  \includegraphics[width=\linewidth]{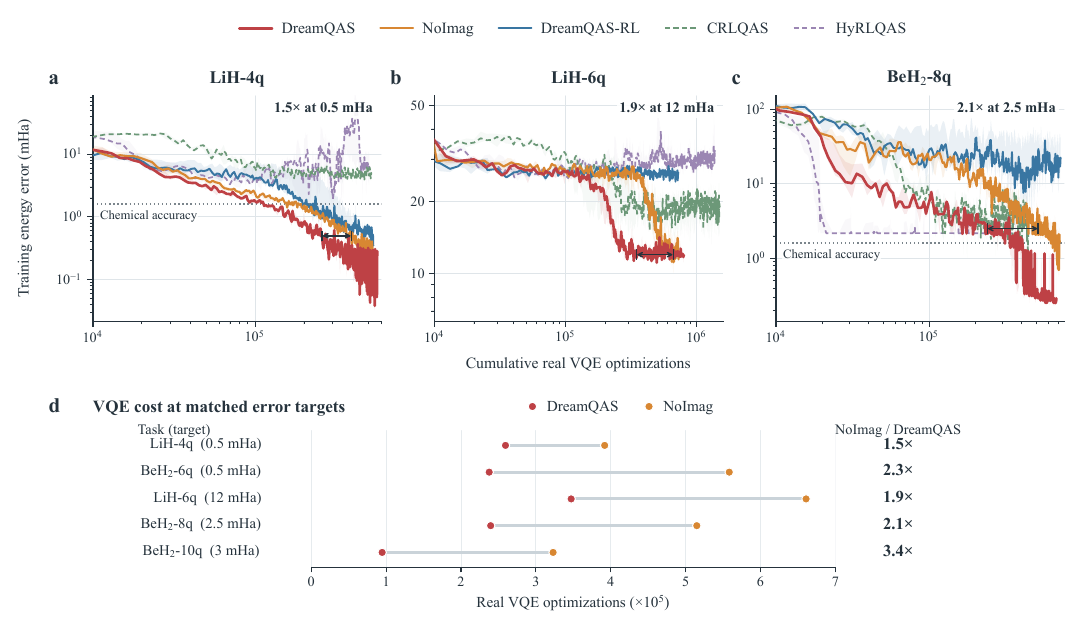}
  \caption{\textbf{Imagination reduces VQE cost at matched error targets.} \textbf{a--c}, Seed median and interquartile range of episode-best error, averaged over the preceding 100 episodes. Arrows compare DreamQAS and NoImag at matched targets. \textbf{d}, VQE counts for DreamQAS (red) and NoImag (orange) at each task's tightest shared target; ratios are NoImag/DreamQAS. Counts include selective verification.}
  \label{fig:vqe_efficiency}
\end{figure}

At these targets, NoImag requires $1.5$, $2.3$, $1.9$, $2.1$ and $3.4$ times as many VQE optimizations as DreamQAS, respectively (Fig.~\ref{fig:vqe_efficiency}d), even though DreamQAS's count includes selective verification. On most tasks, the saving appears as circuits are refined to low errors: at coarse error targets, the ratios range from $0.7$ to $1.1$ on four tasks, so DreamQAS does not consistently require fewer optimizations early in training (Supplementary Table~\ref{tab:supp_gap_noimag}). This is consistent with the design of Section~\ref{sec:dreamqas_overview}: imagined updates are enabled only after the warm-up period and while the ranking check is passed, so imagination cannot contribute at the start of training. BeH$_2$-8q is the exception, with a ratio of $3.1$ already at $20\mha$. DreamQAS-RL needs $1.8$ and $1.6$ times as many optimizations as DreamQAS at the same targets on LiH-4q and BeH$_2$-6q and does not reach them on the other three tasks (Supplementary Table~\ref{tab:supp_gap_noimag}). NoImag remains the primary comparison because it differs from DreamQAS only in the imagined-learning loop; the curves of CRLQAS and HyRLQAS are shown for context.

\subsubsection{Quality of the learned circuit-building policies.}

We next compare the policies obtained after the same number of training episodes. This comparison matches training length rather than VQE cost, since the number of VQE optimizations per episode differs between methods. Each trained policy is frozen and builds circuits in 100 new episodes per seed. Because any circuit produced along the way can be kept as the output, we record the lowest energy error reached in each episode, average it for each seed, and report the mean and sample standard deviation over the five seeds (Table~\ref{tab:main_quality}). DreamQAS has the lowest mean error among the RL-QAS methods on all five tasks and reaches chemical accuracy ($1.6\mha$) on all of them except LiH-6q. The internal controls show where this gain comes from: NoImag improves on DreamQAS-RL on four tasks but not on BeH$_2$-6q, whereas adding imagination improves on NoImag on all five. When only the final circuit of each episode is scored, DreamQAS remains well below CRLQAS, HyRLQAS and DreamQAS-RL on every task and below NoImag on all but BeH$_2$-8q, where the two are close ($45.1$ and $44.6\mha$; Supplementary Table~\ref{tab:supp_final_circuit}). We further evaluated two additional tasks: BeH$_2$-12q, which enlarges the gate set to 168 choices, and H$_2$O-8q, a different molecule. DreamQAS also has the lowest mean final-circuit error among the RL-QAS methods on both extension tasks. Under episode-best evaluation, it ranks first on BeH$_2$-12q but behind HyRLQAS on H$_2$O-8q (Supplementary Table~\ref{tab:supp_final_circuit}).

\begin{table}[!ht]
\centering
\fontsize{8}{10.4}\selectfont
\setlength{\tabcolsep}{2.5pt}
\renewcommand{\arraystretch}{1.16}
\begin{tabular}{@{}>{\raggedright\arraybackslash}p{0.22\linewidth}*{5}{>{\centering\arraybackslash}p{\dimexpr0.156\linewidth-2\tabcolsep\relax}}@{}}
\toprule
\textbf{Method} & \textbf{LiH-4q} & \textbf{BeH$_2$-6q} & \textbf{LiH-6q} & \textbf{BeH$_2$-8q} & \textbf{BeH$_2$-10q}\\
\midrule
\multicolumn{6}{@{}l}{\emph{RL-QAS methods: common training-episode budget}}\\[2pt]
DreamQAS & $\mathbf{0.053\!\pm\!0.024}$ & $\mathbf{0.058\!\pm\!0.000}$ & $\mathbf{11.4\!\pm\!0.47}$ & $\mathbf{0.256\!\pm\!0.033}$ & $\mathbf{0.528\!\pm\!0.327}$\\
\hspace*{1em}NoImag & $\underline{0.221\!\pm\!0.062}$ & $0.328\!\pm\!0.120$ & $\underline{13.6\!\pm\!3.5}$ & $\underline{1.929\!\pm\!1.766}$ & $1.399\!\pm\!0.613$\\
\hspace*{1em}DreamQAS-RL & $0.391\!\pm\!0.074$ & $\underline{0.098\!\pm\!0.061}$ & $26.3\!\pm\!1.1$ & $23.1\!\pm\!17.3$ & $5.054\!\pm\!3.586$\\
CRLQAS~\cite{patel2024curriculum} & $4.730\!\pm\!1.300$ & $1.780\!\pm\!1.300$ & $17.3\!\pm\!3.9$ & $3.360\!\pm\!3.173$ & $1.026\!\pm\!0.288$\\
HyRLQAS~\cite{niu2025hybrid} & $16.5\!\pm\!24$ & $309\!\pm\!690$ & $215\!\pm\!410$ & $2.175\!\pm\!0.000$ & $\underline{0.897\!\pm\!0.000}$\\
\addlinespace[4pt]
\midrule
\multicolumn{6}{@{}l}{\emph{Other QAS methods: method-specific search protocols}}\\[2pt]
GQE~\cite{nakaji2024generative} & $128\!\pm\!79$ & $235\!\pm\!220$ & $108\!\pm\!45$ & $261\!\pm\!130$ & $345\!\pm\!400$\\
TFQAS~\cite{he2024training} & $32.7\!\pm\!1.3$ & $20.3\!\pm\!4.0$ & $36.0\!\pm\!0.62$ & $56.7\!\pm\!10$ & $77.9\!\pm\!5.4$\\
QuantumDARTS~\cite{wu2023quantumdarts} & $5.140\!\pm\!1.550$ & $1.270\!\pm\!2.600$ & $15.2\!\pm\!9.41$ & $2.480\!\pm\!0.576$ & $0.897\!\pm\!0.000$\\
\bottomrule
\end{tabular}
\caption{\textbf{Final energy error across five molecular tasks.} Mean $\pm$ sample SD over five runs (mHa). Each delivered circuit is scored by the lowest error reached in a complete episode; RL-QAS values average 100 episodes per run. An SD of 0.000 means that all runs reached the same energy. Indented rows are the internal controls (Section~\ref{sec:dreamqas_overview}). Bold and underlining mark the lowest and second-lowest RL-QAS means.}
\label{tab:main_quality}
\end{table}

The lower block of Table~\ref{tab:main_quality} places these results alongside the three QAS methods from other families. Because these methods search and evaluate circuits in different ways, this block compares final circuit quality only, not cost. {\color{black}DreamQAS has} the lowest error of all compared methods on all five tasks. Under the method-specific search and optimization protocols listed in Supplementary Table~\ref{tab:supp_methods}, DreamQAS has the lowest reported mean error on these tasks. Because the optimization budgets differ, this comparison does not isolate architecture quality from post-search optimization effort. Taken together, the comparisons in this section show that imagination reduces the VQE optimizations needed to reach low errors and yields better policies for the same training length.

\FloatBarrier
\subsection{Exact gate rules and multi-step policy training make predicted feedback useful}\label{sec:deployment_mechanism}

\begin{figure}[!ht]
  \centering
  \includegraphics[width=\linewidth]{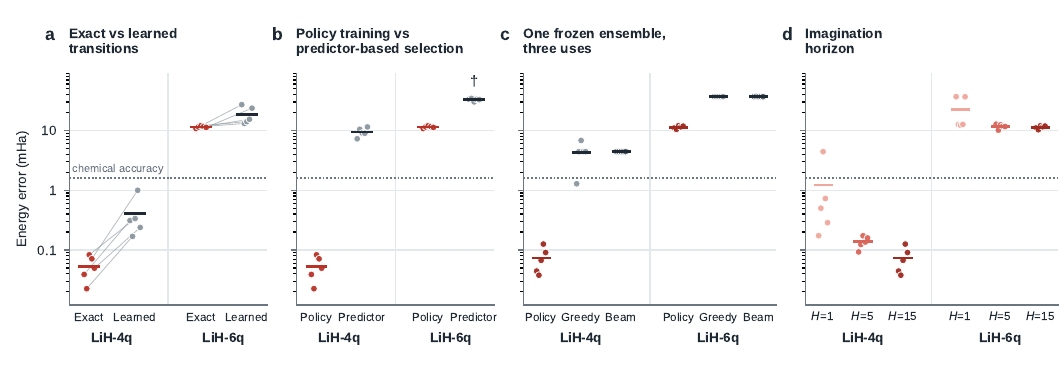}
  % \caption{\textbf{Exact gate rules and multi-step policy training make predicted feedback useful.} Each point is one training seed, averaged over 100 evaluation episodes (greedy and beam selection, which are deterministic, produce one circuit per seed); horizontal bars are means over five seeds, and the dotted line marks chemical accuracy ($1.6\mha$). Red shades denote DreamQAS and its horizon variants; grey denotes alternatives. \textbf{a}, DreamQAS with exact gate rules (Exact) versus a DreamerV3-style model that learns the circuit transitions (Learned); lines connect seeds with the same index. \textbf{b}, DreamQAS, which uses predictions to train its policy (Policy), versus Predictor-QAS, which uses a learned predictor to select circuits (Predictor); DreamQAS points are those in \textbf{a}. \textbf{c}, One frozen ensemble, from a DreamQAS checkpoint trained without selective verification, used with the policy trained alongside it or to select gates greedily or by beam search. \textbf{d}, Imagination horizons $H$ with about the same number of imagined steps per update; the $H=15$ runs are those in \textbf{c}, and $H=1$ and $H=5$ are trained separately. All checkpoints are taken after 15,000 training episodes, except LiH-6q Predictor-QAS ($\dagger$; 30,000 episodes).}
  {\color{black}
  \caption{\textbf{Exact gate rules and multi-step policy training make predicted feedback useful.} Each point is one seed, averaged over 100 evaluation episodes (greedy and beam selection give one deterministic circuit per seed); bars are means over five seeds, and the dotted line marks chemical accuracy. Red shades, DreamQAS and its horizon variants; grey, alternatives. \textbf{a}, Exact: DreamQAS with exact gate rules; Learned: a DreamerV3-style model that learns circuit transitions; lines join matching seeds. \textbf{b}, Policy: DreamQAS, which trains its policy on predictions (points as in \textbf{a}); Predictor: Predictor-QAS, which uses predictions to select circuits. \textbf{c}, One frozen ensemble, from a run without selective verification, used by its trained policy or for greedy or beam selection. \textbf{d}, Horizons $H$ with approximately matched numbers of imagined steps per update; $H=15$ as in \textbf{c}. Checkpoints are taken after 15,000 training episodes ($\dagger$, LiH-6q Predictor-QAS: 30,000).}
  \label{fig:design_controls}
  }
\end{figure}

%Three design choices in Section~\ref{sec:dreamqas_overview} distinguish how DreamQAS uses predicted feedback: imagined circuits are built with the exact rule $T$ rather than a learned model of how circuits change; predictions enter only through the imagined term of Eq.~\eqref{eq:res_policy_loss}, which trains the policy, rather than being used to select circuits; and each imagined sequence runs for up to $H$ gate additions, so that Eq.~\eqref{eq:res_imagined_return} can credit earlier choices with later predicted improvements. 
Three choices distinguish DreamQAS: preserving exact circuit transitions, using predictions for policy learning rather than direct selection, and learning from multi-step imagined trajectories.
We test whether each choice matters by changing it on LiH-4q and LiH-6q and comparing the energy errors of the resulting circuits (Fig.~\ref{fig:design_controls}). The learned-transition comparison replaces the whole world model; for the comparison of how predictions are used, which also changes the model, we add a second control that isolates the use.

\subsubsection{Preserving exact circuit transitions.}

We first ask whether imagination needs the exact gate rules, or whether a learned model of how circuits change would serve as well. 
We replace {\color{black}the exact gate rule and the feedback ensemble} with a DreamerV3-style world model~\cite{hafner2025mastering}, which learns circuit transitions in a compressed representation and predicts the score with its own prediction heads. The VQE optimizer and training-episode budget are unchanged; this is an alternative world-model design rather than a single-component replacement. With learned transitions, the mean error rises from $0.053$ to $0.411\mha$ on LiH-4q and from $11.4$ to $18.7\mha$ on LiH-6q, and the learned model is worse in all five seed-matched comparisons on both tasks (Fig.~\ref{fig:design_controls}a). For the world-model design tested here, learning transitions that are already known therefore brings no benefit and lowers the quality of the final circuits.

\subsubsection{Using predictions for policy learning rather than direct selection.}

We next ask whether predictions are more useful for training the policy or for selecting circuits directly. Predictor-QAS takes the second route, as in earlier predictor-based QAS~\cite{zhang2021neural,deng2023progressive,soloviev2024trainability}: it trains a predictor on VQE-evaluated circuits during search and then uses it to pick the most promising of 512 newly generated circuits. Its circuits have mean errors of $9.46$ and $32.8\mha$ on LiH-4q and LiH-6q, compared with $0.053$ and $11.4\mha$ for DreamQAS (Fig.~\ref{fig:design_controls}b), even though the LiH-6q predictor was trained for twice as many episodes. This comparison, however, changes both the model and the way its predictions are used. To isolate the use, we take one frozen ensemble, from a DreamQAS checkpoint trained without selective verification, and compare three ways of using it: the policy trained alongside it through imagination, 
greedy selection, at each step, of the gate with the {\color{black}best predicted score after the disagreement penalty} (Eq.~\eqref{eq:res_imagined_reward}), and beam search that keeps the ten most promising partial circuits (Fig.~\ref{fig:design_controls}c). The trained policy gives the lowest error on both tasks. On LiH-4q, it reaches $0.073\mha$, within chemical accuracy, whereas greedy and beam selection give $4.28$ and $4.43\mha$; on LiH-6q, the corresponding errors are $11.4$, $36.9$ and $36.9\mha$. For the tested tasks and selection schemes, the same predictions are thus more useful for training a policy than for choosing circuits directly.

\subsubsection{Learning from multiple imagined gate additions.}

A gate addition may become useful only after subsequent gates are added. We therefore {\color{black}set the imagination horizon $H$ to $1$, $5$ or $15$} (Fig.~\ref{fig:design_controls}d). With $H=1$, each imagined {\color{black}sequence} contains a single predicted reward, so a later predicted improvement cannot credit earlier choices within an imagined sequence. To keep the amount of imagined experience comparable, the one- and five-step runs retain about the same number of imagined steps per policy update as the fifteen-step runs, and all three settings are trained without selective verification. This comparison therefore tests whether longer sequences help when the number of imagined training steps is approximately matched. On both tasks, the five- and fifteen-step settings give lower mean errors than the one-step setting: on LiH-4q, the error falls from $1.22\mha$ with one step to $0.137$ and $0.073\mha$ with five and fifteen steps, and on LiH-6q it falls from $22.2$ to $11.8$ and $11.4\mha$. On LiH-6q, the one-step mean reflects two of the five runs, which end near $36.8\mha$; the other three, at $12.3$--$12.7\mha$, fall within the range of the multi-step runs. The difference between five and fifteen steps is much smaller than the gain of either over one step. These results support learning across several consecutive gate additions, but they do not identify an optimal horizon. 

In each of the three comparisons, the lowest mean errors on both tasks come from the choices DreamQAS makes: exact gate rules, predictions used to train the policy, and imagined sequences of several gate additions. Repeating the second and third comparisons on BeH$_2$-8q gives the same ordering (Supplementary Table~\ref{tab:supp_design_controls}): the trained policy reaches $0.26\mha$, compared with $1.40\mha$ for greedy selection and beam search and $63.4\mha$ for Predictor-QAS, and the one- and five-step settings reach $1.04$ and $0.36\mha$.

\FloatBarrier
\subsection{The feedback ensemble learns useful circuit rankings and identifies unreliable predictions}\label{sec:utility_risk}

%Section~\ref{sec:dreamqas_overview} showed that imagined updates are guided by how the ensemble ranks circuits, and that its disagreement $\sigma_t$ is used to limit unreliable predictions. 
Imagined learning relies on predicted differences in circuit quality rather than direct VQE evaluations.
We therefore test whether the ensemble learns useful next-gate rankings and whether disagreement identifies inaccurate feedback. {\color{black}Before these tests, we check the learning target itself.} The frontier-relative rewards of real gate additions agree closely with rewards computed relative to the exact ground-state energy: late-training Spearman correlations are $0.973$--$1.000$, with almost complete agreement in reward sign (Supplementary Table~\ref{tab:supp_reward_audit}). This supports the observed frontier as a practical learning reference, although the experimental curriculum uses an exact-energy-anchored reference.

%We test ranking on partial circuits built by the frozen policy (Fig.~\ref{fig:action_ranking} and Table~\ref{tab:action_utility}) and reliability on circuits checked by VQE during training (Fig.~\ref{fig:risk_coverage}).

\begin{figure}[!ht]
\centering
\begin{minipage}[c]{0.49\linewidth}
  \centering
  \includegraphics[width=\linewidth]{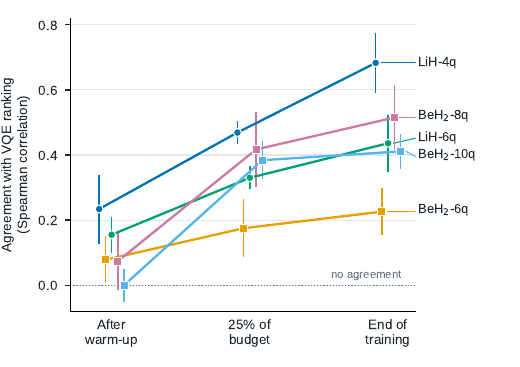}
  \caption{\textbf{The feedback ensemble learns to rank next gates.} Rank correlation between the ensemble's and VQE's orderings of up to ten possible next gates for new partial circuits. Mean $\pm$ SD over five seeds; checkpoint budgets are listed in Supplementary Table~\ref{tab:supp_action_utility}.}
  \label{fig:action_ranking}
\end{minipage}\hfill
\begin{minipage}[c]{0.48\linewidth}
  \centering
  \setlength{\tabcolsep}{3pt}\renewcommand{\arraystretch}{1.6}
  \begin{tabular*}{\linewidth}{@{\extracolsep{\fill}}lcc@{}}
  \toprule
  \textbf{Task} & \textbf{Regret} $\downarrow$ & \shortstack{\textbf{No worse}\\\textbf{than random} $\uparrow$}\\
  \midrule
  LiH-4q & $0.026\!\pm\!0.046$ & $0.942\!\pm\!0.040$\\
  BeH$_2$-6q & $0.152\!\pm\!0.056$ & $0.649\!\pm\!0.106$\\
  LiH-6q & $0.054\!\pm\!0.048$ & $0.707\!\pm\!0.078$\\
  \bottomrule
  \end{tabular*}
  \vspace{2pt}
  \captionof{table}{\textbf{Gate ranked first by the ensemble.} Final checkpoint; mean $\pm$ SD over five seeds. A random choice scores at least 0.5 in the last column. BeH$_2$-8q and BeH$_2$-10q, with many tied gates, are in Supplementary Table~\ref{tab:supp_action_utility}.}
  \label{tab:action_utility}
\end{minipage}
\end{figure}

\subsubsection{Learning to rank candidate gates}

To test ranking, we freeze the policy at early, intermediate and final diagnostic checkpoints and use it to build partial circuits. For each circuit, the ensemble ranks up to ten possible next gates, and VQE evaluates each of them. The {\color{black}rank correlation} between the two orderings rises during training on all five tasks, by $0.35$ on average from the first checkpoint to the last, and ends between $0.23$ on BeH$_2$-6q and $0.68$ on LiH-4q (Fig.~\ref{fig:action_ranking}). {\color{black}At the final checkpoint, we evaluate the gate ranked first by the ensemble with two measures (Table~\ref{tab:action_utility}): its regret, the shortfall from the best tested gate as a fraction of the range among the tested gates, and the probability that it is no worse than a randomly chosen gate from the same set, which is at least $0.5$ for a random choice. On LiH-4q, BeH$_2$-6q and LiH-6q, the regret is $0.026$, $0.152$ and $0.054$, and the probability is $0.94$, $0.65$ and $0.71$.} On BeH$_2$-8q and BeH$_2$-10q, many tested gates reach the same VQE energy, so these two measures are reported in Supplementary Table~\ref{tab:supp_action_utility}. The ensemble thus learns an informative but far from exact ordering of next gates, and BeH$_2$-6q, the task on which its ordering agrees least with VQE, is also the one on which its first choice is weakest.

\subsubsection{Disagreement and prediction reliability}

\begin{figure}[!ht]
  \centering
  \includegraphics[width=\linewidth]{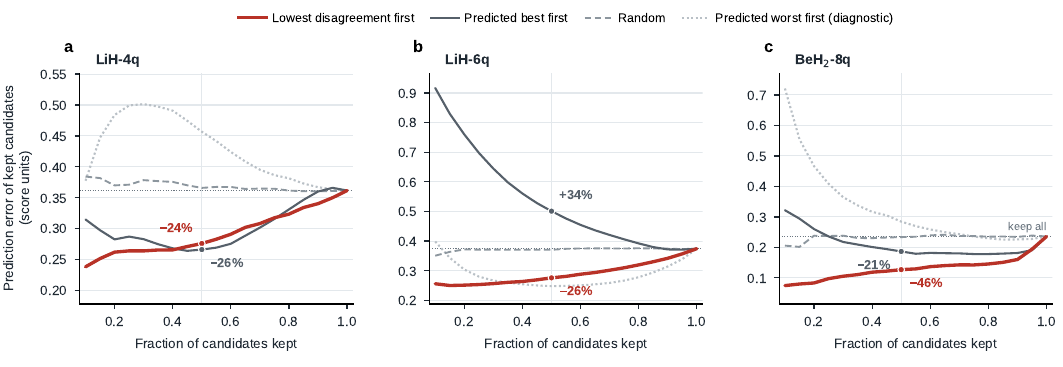}
  \caption{\textbf{Ensemble disagreement flags unreliable predictions.} Candidates sent to VQE during training for their predicted quality are kept in four different orders. The vertical axis gives the mean prediction error of the kept candidates, in units of the {\color{black}frontier-relative score} of Eq.~\eqref{eq:res_score} rather than mHa; curves are means over five seeds. Percentages give the change at half coverage relative to keeping all candidates (dotted line).}
  \label{fig:risk_coverage}
\end{figure}

{\color{black}To test whether disagreement identifies inaccurate predictions, we} examine verified candidates selected because of predicted quality, including candidates also selected for disagreement, on LiH-4q, LiH-6q and BeH$_2$-8q. We order candidates by increasing ensemble disagreement and calculate mean absolute prediction error as progressively larger fractions are retained. Errors are measured in units of the frontier-relative score, not mHa. We compare this ordering with random ordering and with ordering by predicted {\color{black}score} (Fig.~\ref{fig:risk_coverage}).

%If disagreement tracks reliability, this error should fall as fewer circuits are kept. We compare this order with a random order and with an order by predicted score $\bar y$, best first, which asks whether disagreement adds information beyond how promising a prediction looks. The reverse order, worst predicted score first, is shown as a diagnostic.

%Keeping circuits with low disagreement lowers the prediction error on all three tasks (Fig.~\ref{fig:risk_coverage}). 
When half of the circuits are kept, the error falls by $24\%$, $26\%$ and $46\%$ on LiH-4q, LiH-6q and BeH$_2$-8q relative to keeping all of them, whereas a random order leaves it almost unchanged. Averaged over all fractions kept, the disagreement order gives a lower error than the random order in each of the 15 combinations of task and seed (Supplementary Table~\ref{tab:supp_risk_coverage}). Ordering by predicted score, by contrast, helps on two tasks and hurts on the third: keeping the most promising half lowers the error by $26\%$ on LiH-4q {\color{black}and} $21\%$ on BeH$_2$-8q but raises it by $34\%$ on LiH-6q. Averaged over all fractions kept, disagreement gives a lower error than this order in 13 of the 15 combinations. On LiH-6q, the reverse order, which keeps the least promising circuits first, does as well as or slightly better than disagreement. A promising prediction is thus not necessarily a reliable one: disagreement identifies unreliable predictions on all three tasks, and does so better than the predicted score on two of them.

\subsubsection{Contribution of {\color{black}individual components}}

Finally, on all five tasks, we change one component at a time: we remove selective verification, remove the disagreement-based controls, or replace density weighting{\color{black}, which upweights rarely observed scores when training the ensemble,} with an earlier fixed weighting of the scores (Supplementary Table~\ref{tab:supp_ablation}). With any single component changed, imagination still gives a lower mean error than NoImag on all five tasks. On the three smaller tasks, the largest change is on LiH-4q, where replacing density weighting raises the error from $0.053$ to $0.125\mha$; the small reductions on LiH-6q, to $11.1\mha$ with the fixed weighting and $11.2\mha$ without the disagreement-based controls (versus $11.4\mha$), are within the spread across seeds. On BeH$_2$-8q and BeH$_2$-10q, every variant has a higher mean error than DreamQAS. The increase is at most $0.06\mha$ on BeH$_2$-8q and is clearest on BeH$_2$-10q, where replacing density weighting raises the error from $0.53$ to $0.87\mha$ and removing the disagreement-based controls or selective verification raises it to $0.66$ and $0.67\mha$; these differences are comparable to the spread across seeds. The benefit of imagination therefore does not depend on any single {\color{black}component}, although the full method has the lowest mean error on the two largest tasks.

\FloatBarrier
\subsection{DreamQAS retains the lowest mean error under simulated noise}\label{sec:noise}

All experiments so far used noise-free VQE energies. We last ask whether DreamQAS remains useful when every energy it receives is noisy. On LiH-4q, we simulate gate and readout errors at two levels, with rates resembling those of two IBM devices (Methods). These are channel simulations rather than hardware runs, and they omit finite-shot sampling. DreamQAS, NoImag, DreamQAS-RL, CRLQAS and HyRLQAS are trained for the same number of episodes with five seeds each, and their final policies select circuits using the same noisy energies. We report the error of each selected circuit from its noisy energy (Table~\ref{tab:noise}); this error therefore reflects both how well the circuit approximates the ground state and how strongly noise shifts its energy, and it is not comparable with the noise-free errors in Table~\ref{tab:main_quality}.

\begin{table}[!ht]
\centering
\fontsize{8}{10.4}\selectfont
\setlength{\tabcolsep}{2.5pt}
\renewcommand{\arraystretch}{1.24}
\begin{tabular*}{\linewidth}{@{\extracolsep{\fill}}lccccc@{}}
\toprule
\textbf{Noise profile} & \textbf{DreamQAS} & \textbf{NoImag} & \shortstack{\textbf{DreamQAS-}\\\textbf{RL}} & \textbf{CRLQAS} & \textbf{HyRLQAS}\\
\midrule
Low noise & $\mathbf{14.64\!\pm\!0.43}$ & $14.95\!\pm\!0.16$ & $22.48\!\pm\!2.30$ & $17.80\!\pm\!1.04$ & $154.88\!\pm\!95.44$\\
High noise & $\mathbf{45.32\!\pm\!0.16}$ & $45.65\!\pm\!0.33$ & $47.07\!\pm\!1.87$ & $49.36\!\pm\!1.98$ & $71.08\!\pm\!23.28$\\
\bottomrule
\end{tabular*}
\caption{\textbf{Selected-circuit error on LiH-4q under simulated VQE noise.} Error is the noisy energy of each selected circuit minus the exact ground-state energy, so it includes the shift caused by noise. Mean $\pm$ sample SD over five seeds (mHa); bold marks the lowest mean at each noise level.}
\label{tab:noise}
\end{table}

DreamQAS has the lowest mean error at both noise levels. Its margins over NoImag, $0.31$ and $0.33\mha$, are comparable to the spread across seeds, so the two are not clearly distinguished under noise. The margins over DreamQAS-RL, CRLQAS and HyRLQAS are larger, at least $3.2\mha$ under low noise and $1.8\mha$ under high noise. Under the tested noise, DreamQAS therefore remains the most accurate of the compared methods, but its imagined updates give no clear benefit over NoImag. {\color{black}With one molecular benchmark and two noise profiles, these experiments do not establish general robustness to quantum noise.}  Whether imagined learning can deliver comparable evaluation savings under stochastic hardware feedback remains open.

Overall, the results show that previous VQE feedback can be reused to train a sequential circuit-building policy, reducing the {\color{black}VQE optimizations needed to reach matched error targets} in the noise-free benchmarks studied. {\color{black}The design experiments indicate that the benefit depends on how predictions are used, and the reliability experiments show that ensemble disagreement identifies unreliable predictions.}

\FloatBarrier

%% file: section/discussion.tex
\section{Discussion}\label{sec:discussion}

The repeated optimization of circuit parameters is a major source of computational cost in reinforcement-learning-based quantum architecture search. DreamQAS addresses this limitation by exploiting a distinction between two components of circuit design: the architecture resulting from each gate addition is known exactly, whereas evaluating its quality requires VQE optimization. By preserving the known circuit transitions and learning only the expensive evaluation feedback, DreamQAS enables a circuit-building policy to learn from multi-step imagined experience between real evaluations. Across five molecular ground-state benchmarks, this approach reaches matched {\color{black}energy-error} targets with $1.5$--$3.4$ times fewer VQE optimizations than its non-imagining counterpart (Fig.~\ref{fig:vqe_efficiency}d). It also produces lower mean episode-best energy errors under a common training-episode budget on all seven tasks, including the two extension tasks (Table~\ref{tab:main_quality} and Supplementary Table~\ref{tab:supp_final_circuit}). These findings demonstrate that feedback obtained from previous VQE optimizations can be reused to provide additional policy-learning experience, reducing the need for new circuit evaluations during architecture search.

An important implication is that model-based learning can benefit from incorporating the known structure of the problem rather than approximating every component of the environment. In conventional model-based reinforcement learning, imagined trajectories generally rely on learned predictions of both state transitions and feedback. In sequential QAS, however, the effect of adding a gate to a circuit architecture is deterministic and can be computed exactly. DreamQAS therefore avoids introducing approximation errors into circuit construction and concentrates its learned model on the feedback that would otherwise require VQE optimization. The comparison with a DreamerV3-style learned-transition model supports this design choice, with exact circuit transitions producing lower energy errors on both LiH benchmarks examined (Fig.~\ref{fig:design_controls}a). Although this comparison also changes the world-model architecture and does not isolate transition accuracy alone, it suggests that learning known circuit transitions is unnecessary for effective imagined policy training in the tested settings.

The results further indicate that the value of learned feedback depends not only on its predictive quality, but also on how its predictions are used. Predictor-based QAS methods primarily use learned estimates to rank or select candidate architectures~\cite{zhang2021neural,he2023gnn,deng2023progressive,soloviev2024trainability,martyniuk2025benchmarking}, while planning-based methods evaluate possible future decisions using a model~\cite{schrittwieser2020mastering,rapp2025reinforcement}. DreamQAS instead uses predictions to generate additional training experience for a sequential policy. In the controlled comparison using the same frozen feedback ensemble, the trained policy produces substantially lower energy errors than greedy or beam selection based directly on predicted scores (Fig.~\ref{fig:design_controls}c). {\color{black}This gap does not appear to come from poor single-step choices: on LiH-4q, the gate ranked first by the ensemble is no worse than a random choice with probability $0.94$ (Table~\ref{tab:action_utility}), yet greedy selection ends at $4.28\mha$, compared with $0.073\mha$ for the policy. A more likely difference is how far ahead each use looks: greedy and beam selection rank partial circuits by their predicted score after one more gate, whereas the policy is trained on returns that also credit improvements appearing several gates later.} Moreover, multi-step imagined trajectories improve circuit quality compared with one-step imagination (Fig.~\ref{fig:design_controls}d). These observations are consistent with the possibility that imagined policy learning captures interactions between successive gate choices, including improvements that become apparent only after several operations. However, the experiments do not directly isolate delayed credit assignment from other effects of longer trajectories. The broader finding is that predictions which are imperfect as direct selection criteria can nevertheless provide useful learning signals when incorporated into sequential policy optimization.

This distinction is particularly relevant because the feedback ensemble does not need to predict circuit quality perfectly to support policy learning. Although its agreement with VQE-derived next-gate rankings improves during training, the final rank correlations remain between $0.23$ and $0.68$ across the primary benchmarks (Fig.~\ref{fig:action_ranking}). Nevertheless, policies trained using these predictions outperform the tested direct selection strategies. Because imagined rewards are derived from differences between predicted scores, the learning process depends on the relative quality of successive architectures rather than on exact predictions of their absolute energies. {\color{black}Because only relative quality matters, the} frontier-relative feedback formulation also avoids requiring the exact ground-state energy in the model's learning targets and policy rewards. Its strong agreement with rewards computed using the exact energy (Supplementary Table~\ref{tab:supp_reward_audit}) supports this choice on the tested benchmarks, although the experimental curriculum still uses an exact-energy-anchored reference for convenience. Together, these findings suggest that learning useful relationships between circuit architectures may be more important for imagined policy training than obtaining highly accurate absolute energy estimates.

The ability to learn from imperfect predictions nevertheless introduces a risk: inaccurate feedback can mislead the policy, particularly as it explores architectures that differ from those previously evaluated. DreamQAS {\color{black}is designed to limit this risk with three reliability controls: a ranking check,} ensemble disagreement and selective VQE verification. The reliability experiments show that retaining circuits with lower ensemble disagreement reduces prediction error on all three benchmarks examined (Fig.~\ref{fig:risk_coverage}), indicating that disagreement provides useful information about relative prediction reliability. However, the component-removal experiments show that no individual {\color{black}component} is essential for imagination to outperform NoImag on the five primary tasks (Supplementary Table~\ref{tab:supp_ablation}). Removing disagreement-based controls or selective verification increases the mean error on the two largest primary benchmarks, but these differences are comparable to the variability across seeds. The results therefore support the use of reliability controls without establishing their independent contributions conclusively. Furthermore, ensemble disagreement is not a calibrated estimate of prediction error: predictors may agree on an inaccurate prediction, particularly for architectures outside the observed training distribution. Developing better-calibrated uncertainty estimates and more selective verification strategies may therefore improve the reliability and evaluation efficiency of imagined learning.

Several limitations remain important when interpreting the computational benefits of DreamQAS. First, imagined learning reduces but does not eliminate real VQE optimization. The feedback ensemble requires an initial record of evaluated circuits, and additional VQE evaluations remain necessary as the policy explores new architectures. {\color{black}Consequently, the observed efficiency advantage is most pronounced at lower energy-error targets, while coarse targets reached early in training show no consistent reduction in cost (Supplementary Table~\ref{tab:supp_gap_noimag}).} Selective verification also introduces a direct evaluation cost, accounting for approximately $10$--$12\%$ of VQE optimizations in the timed experiments (Supplementary Fig.~\ref{fig:supp_timing}). Second, the primary efficiency metric counts complete VQE optimizations rather than individual energy evaluations, quantum circuit executions or hardware measurements. Since the computational cost of an optimization depends on factors such as circuit structure, parameter count and optimizer {\color{black}behavior}, the reported reduction in optimization count should not be interpreted as an equivalent reduction in total runtime or quantum hardware cost. Although VQE dominates the measured training time in the state-vector simulations, more detailed resource accounting is needed to establish the computational benefits in larger and hardware-based settings.

A further limitation is that each feedback ensemble and policy is trained independently for a particular molecule and gate set. The additional BeH$_2$-12q and H$_2$O-8q benchmarks provide evidence beyond the five primary tasks, but they do not establish how the approach scales to substantially larger circuit spaces or whether knowledge acquired on one molecular system can be transferred to another. Transferable feedback models could be particularly valuable because the initial VQE evaluations needed to establish a reliable ensemble represent a cost that must currently be paid for every new task. Pretraining on related molecular systems or adapting learned representations across Hamiltonians and gate sets may reduce this initial evaluation burden. Such approaches would also require methods for identifying when previously learned feedback remains reliable under changes in the search space.

The {\color{black}behavior} of imagined learning under realistic quantum noise also remains unresolved. Quantum noise can distort the energy feedback used for circuit search and affect the optimization landscape~\cite{wang2021noise}, motivating noise-aware QAS approaches~\cite{du2022quantum,wang2022quantumnas,patel2024curriculum}. In the simulated-noise experiments on LiH-4q, DreamQAS retains the lowest mean energy error among the compared methods, but its advantage over NoImag is small relative to the variation across seeds (Table~\ref{tab:noise}). These experiments therefore do not establish a clear additional benefit from imagined updates when VQE feedback is noisy. Moreover, the simulations omit finite-shot sampling and several hardware noise processes, leaving open whether the feedback ensemble can learn sufficiently reliable predictions from stochastic measurements on quantum devices. This is an important challenge because measurement noise affects not only the feedback used for policy training but also the data used to train and validate the predictive model. 
Future work should therefore examine imagined learning with finite-shot VQE evaluations, hardware-relevant noise models and adaptive verification strategies that account for both prediction uncertainty and measurement uncertainty. In addition, because resource requirements differ markedly between error-mitigated and error-corrected circuits, future training of DreamQAS can explicitly include targets for both regimes, resulting in optimized architectures commensurate with their respective requirements.

Overall, DreamQAS demonstrates that expensive VQE feedback can be reused to train a sequential circuit-building policy without learning the circuit transitions that are already known exactly. The results {\color{black}show} evaluation savings on the {\color{black}noise-free} molecular benchmarks studied and {\color{black}indicate} that the usefulness of learned feedback depends on how predictions are incorporated into policy learning{\color{black}, while ensemble disagreement helps identify unreliable predictions}. Beyond the particular framework, the central principle is to exploit known structure in a quantum design problem while learning only the component that is expensive to evaluate. This provides a direction for developing more evaluation-efficient quantum architecture search methods and, potentially, other sequential quantum design approaches in which construction rules are known but performance evaluations remain costly.

%% file: section/methods.tex
\section{Methods}\label{sec:methods}

DreamQAS combines VQE-evaluated circuit-building experience with imagined trajectories generated using exact gate additions and learned feedback. We give the mathematical formulation and implementation details here, followed by the experimental protocols used in Results. The molecular search spaces and hyperparameters are specified in Supplementary Tables~\ref{tab:supp_search_spaces} and~\ref{tab:supp_hyperparameters}, and the implementation is publicly available.

\subsection{Circuit construction, VQE feedback and search cost}\label{sec:method_qas_environment}

All reinforcement-learning methods in this work build a parameterized circuit one gate at a time~\cite{ostaszewski2021reinforcement,patel2024curriculum}. {\color{black}At construction step $t$, the state $s_t$ is the ordered sequence of gates added so far, and an action $a_t$ appends one gate from a task-specific set of single-qubit rotations and directed CNOTs. Immediately repeating the same rotation type on the same qubit, or the same CNOT on the same ordered qubit pair, is not allowed. The next state, $s_{t+1}=T(s_t,a_t)$, is therefore known exactly: $T$ is the known gate-addition rule, not a learned model of how circuits change.} After each gate is added, VQE optimizes the continuous parameters of the partial circuit and returns its energy,
\begin{equation}
 E_{\mathrm{VQE}}(s)=\min_{\boldsymbol\vartheta}\,
 \langle 0|U^\dagger(s,\boldsymbol\vartheta)\,\widehat H\,U(s,\boldsymbol\vartheta)|0\rangle ,
 \label{eq:method_vqe_energy}
\end{equation}
where $\widehat H$ is the molecular Hamiltonian and the minimization is carried out approximately by a classical optimizer{\color{black}; we write $E_t=E_{\mathrm{VQE}}(s_t)$}. In DreamQAS and its variants, previously optimized angles are kept within an episode and each new rotation starts at zero. Following HamQASBench~\cite{niu2026hamqasbench}, tasks with four or six qubits use COBYLA with at most 1,000 iterations, and tasks with eight or more qubits use ROTOSOLVE with three sweeps.

We measure search cost in VQE optimizations: each evaluation of a partial circuit, including the optimization of its parameters, counts as one. Optimizations used for selective verification (below) are included, whereas those used for frozen evaluation and diagnostics are not; CRLQAS and HyRLQAS are counted in the same way. A training episode ends at the maximum depth or when its energy crosses a curriculum threshold. The curriculum follows CRLQAS~\cite{patel2024curriculum} and uses a fixed reference energy below all energies reached during search; for convenience we set it $2.28$~Ha below {\color{black}the exact ground-state energy} $E_0$. Because each curriculum update resets the stopping threshold relative to the best energy observed so far, the reference energy affects only the threshold used before the first update. Frozen evaluation ignores the curriculum and runs each episode to the maximum depth.

\subsection{Learning targets relative to the best energy found so far}\label{sec:method_targets}

The feedback ensemble learns neither raw energies, whose scale differs between molecules, nor errors relative to $E_0$, which is unavailable for most molecules of interest. {\color{black}Instead, it learns a score relative to the frontier $F$, the lowest energy observed so far in real episodes and selective verification,
\begin{equation}
 y=S_F(E)=\operatorname{sign}(E-F)\,\log_{10}\!\left(1+\frac{|E-F|}{m}\right),
 \label{eq:res_score}
\end{equation}
with scale $m=0.1\mha$.} The frontier $F_k$ is adopted at the $k$th ensemble refresh and held fixed until the next, so the model is trained against a consistent reference between updates. Because $S_F$ is strictly increasing in $E$, moving the frontier changes the scores but not the ranking of any set of energies, and the logarithm compresses large deviations. The replay buffer stores the evaluated gate sequences and raw energies, so all recorded partial circuits are rescored whenever the adopted frontier changes and earlier data remain usable. With $y_t=S_{F_k}(E_t)$, the real reward is $r_{t+1}=y_t-y_{t+1}$, so a step to a circuit with lower energy receives a positive reward. The exact ground-state energy is used to set the curriculum reference and to report evaluation errors, but not to construct learning targets or rewards.

\subsection{Feedback ensemble}\label{sec:method_feedback_ensemble}

The ensemble has $K=3$ members, each consisting of a gated recurrent unit (GRU) that reads the gate sequence in order, a prediction head and a fixed, randomly initialized prior network added to the output~\cite{osband2018randomized}. Each member is trained on a different random subset of the data~\cite{lakshminarayanan2017simple}, which keeps the members diverse as the data change. The mean $\bar y_t$ and standard deviation $\sigma_t$ of the member predictions are the predicted score and the disagreement used in Eqs.~\eqref{eq:res_imagined_reward}--\eqref{eq:res_policy_loss}. The policy receives the average of the members' recurrent representations, $\bar{\mathbf h}_t$, with gradients stopped, so policy updates do not change the ensemble.

The ensemble is trained by weighted Huber regression~\cite{huber1992robust} on the scores of all partial circuits in replayed episodes. Training batches mix four sampling streams: an elite stream that draws whole episodes ranked by their lowest observed energy, a stream favoring recent and high-quality episodes, a stream stratified by score and a uniform stream. Inverse-density weights~\cite{yang2021delving}{\color{black}, which we refer to as density weighting,} increase the contribution of rarely observed score values. At periodic refreshes, the frontier is updated, all recorded scores are recomputed and the ensemble is retrained; the sampling proportions and update schedule are given in Supplementary Table~\ref{tab:supp_hyperparameters}.

\subsection{Policy training with real and imagined experience}\label{sec:method_imagined_learning}

The policy $\pi_\theta(a_t\mid s_t)${\color{black}, with parameters $\theta$,} is implemented as a categorical distribution over the gate set, conditioned on $\bar{\mathbf h}_t$, with illegal gates masked. Each policy iteration collects four real episodes. Their rewards give discounted returns, standardized within the batch, which weight the REINFORCE loss $\mathcal L_{\mathrm{real}}$ together with an entropy bonus~\cite{williams1992simple}; no learned critic is used.

For imagined experience, DreamQAS samples recent recorded episodes, cuts each at a random point and extends the resulting partial circuit by at most $H=15$ gate additions without VQE. At each step the policy samples a legal gate, the exact rule $T$ builds the next circuit and the ensemble predicts its score. {\color{black}The imagined reward uses a pessimistic version of the predicted score,
\begin{equation}
 \widehat\Phi_t=-\left(\bar y_t+\beta\,\sigma_t\right),\qquad
 \widehat r_{t+1}=\widehat\Phi_{t+1}-\widehat\Phi_t ,
 \label{eq:res_imagined_reward}
\end{equation}
where $\beta>0$ sets how strongly disagreement counts against a circuit, so a circuit on which the predictors disagree is treated as worse than its mean prediction. Returns are accumulated backwards along the imagined sequence,
\begin{equation}
 \widehat G_t=z_t\,\widehat r_{t+1}+\gamma\lambda\,\widehat G_{t+1},
 \label{eq:res_imagined_return}
\end{equation}
starting} from $\widehat G_L=0$, where $L\leq H$ is the generated sequence length; $\gamma$ discounts future feedback and $\lambda$ further attenuates distant imagined rewards. {\color{black}The indicator $z_t$ equals one for an accepted imagined transition and zero for a rejected one (see the disagreement controls below).} Subtracting the batch mean return at each imagined step gives $\widehat A_t$; no learned value estimate is used. One optimizer step then minimizes{\color{black}
\begin{equation}
 \mathcal L_{\mathrm{policy}}
 =\mathcal L_{\mathrm{real}}
 -\omega\, I_{\mathrm{rank}}\,
 \frac{\sum_t z_t\,\widehat A_t\log\pi_\theta(a_t\mid s_t)}{\sum_t z_t},
 \label{eq:res_policy_loss}
\end{equation}
}in which $\omega$ weights imagined experience and $I_{\mathrm{rank}}$ is set by the ranking check below; the imagined term is {\color{black}defined as zero when $I_{\mathrm{rank}}=0$ or when no imagined transitions are accepted ($\sum_t z_t=0$)}. The number of sampled episodes and the values of $\gamma$, $\lambda$ and $\omega$ are listed in Supplementary Table~\ref{tab:supp_hyperparameters}.

\subsection{Controls on unreliable predictions}\label{sec:method_reliability}

The ranking check decides whether imagined experience enters the policy update. Before each ensemble update, the ensemble scores the partial circuits recorded in the 200 most recent episodes, and we estimate from randomly sampled pairs with different VQE energies the fraction that it orders correctly,
\begin{equation}
 \mathrm{Acc}_{\mathrm{pair}}
 =\Pr_{(u,v)}\!\left[(\bar y_u-\bar y_v)(E_u-E_v)>0\right].
 \label{eq:method_ranking_gate}
\end{equation}
After a warm-up of 500 real episodes, $I_{\mathrm{rank}}=1$ while $\mathrm{Acc}_{\mathrm{pair}}\geq 0.70$ and $I_{\mathrm{rank}}=0$ otherwise. This statistic is a heuristic for switching imagination on and off, not an estimate of accuracy on new circuits.

The disagreement controls set $\beta=1$ in Eq.~\eqref{eq:res_imagined_reward}. In addition, an imagined sequence stops when $\sigma_t$ exceeds $0.6$, when no legal gate remains or when $H$ steps have been taken; the step that exceeds the disagreement threshold is rejected and contributes no policy gradient ($z_t=0$).

Selective verification runs every 20 policy iterations. DreamQAS generates imagined circuits from separately sampled recorded partial circuits and selects the five with the lowest predicted score and the five with the largest disagreement. Each selected circuit is rebuilt from the empty circuit with a VQE evaluation after every gate, and the resulting energies are added to the replay buffer, in the spirit of dataset aggregation~\cite{ross2011reduction}. These checks run throughout training, including before imagined updates are switched on, and are included in the counts of VQE optimizations.

\subsection{Tasks, compared methods and evaluation}\label{sec:method_evaluation}

The five primary tasks comprise LiH with four and six qubits and BeH$_2$ with six, eight and ten qubits; BeH$_2$ with 12 qubits and H$_2$O with 8 qubits serve as extension tasks. The BeH$_2$ task specifications follow HamQASBench~\cite{niu2026hamqasbench}, whereas the LiH and H$_2$O molecular setups follow earlier CRLQAS and HyRLQAS studies~\cite{patel2024curriculum,niu2025hybrid} (Supplementary Tables~\ref{tab:supp_molecules}--\ref{tab:supp_search_spaces}). The maximum depth is 40 gates for LiH-4q and 50 gates for the other tasks. Every reinforcement-learning method is trained for 15,000 episodes with five seeds per task.

NoImag keeps the feedback ensemble and the policy's recurrent representation but makes no imagined updates{\color{black} ($\omega=0$ in Eq.~\eqref{eq:res_policy_loss})}, and therefore performs no selective verification, which acts on imagined circuits. DreamQAS-RL applies REINFORCE directly to an encoding of the circuit structure, with real feedback only and no ensemble. CRLQAS~\cite{patel2024curriculum} and HyRLQAS~\cite{niu2025hybrid} keep their original reward rules and curricula; for convenience these curricula are anchored at $E_0$, although an energy known to lie below $E_0$ could serve instead. GQE~\cite{nakaji2024generative}, TFQAS~\cite{he2024training} and QuantumDARTS~\cite{wu2023quantumdarts} follow their own search protocols (Supplementary Table~\ref{tab:supp_methods}) and are compared on final circuit quality only.

For the primary comparisons, reported policies are the checkpoints reached after the fixed budget of 15,000 training episodes; checkpoints are not selected by performance. Each frozen policy samples 100 new episodes per seed. The result of an episode is the lowest error $|E-E_0|$, in mHa, among its VQE-evaluated partial circuits; the 100 results are averaged per seed, and tables report the mean and sample standard deviation over seeds. Errors of the final circuit alone are reported in Supplementary Table~\ref{tab:supp_final_circuit}.

For the efficiency curves, the episode results of each training run are averaged over the most recent 100 episodes and recorded every 50 episodes against the cumulative number of VQE optimizations; the curves show the median and interquartile range over seeds. For each task, the error target is the tightest value in a fixed set that both DreamQAS and NoImag reach. A method reaches a target when its median curve stays at or below it for three consecutive recorded points, and its cost is the cumulative VQE count at the first of these points.

The training seed is the experimental unit. Timing runs use three seeds per method and are reported only as within-run shares (Supplementary Fig.~\ref{fig:supp_timing}). Where relevant, comparisons report the number of seed-matched pairs favoring one variant; no significance tests are applied to the main-text comparisons.

\subsection{Design-control experiments}\label{sec:method_design_controls}

The learned-transition alternative follows DreamerV3~\cite{hafner2025mastering}. A recurrent state-space model reads the encoding of the circuit structure and learns deterministic and categorical latent states, together with a decoder that reconstructs the structure. During imagination, the latent state advances through the learned prior without access to the true circuit, and the model's own $K$ prediction heads, attached to the latent state, predict the score of Eq.~\eqref{eq:res_score}. Gate legality is still enforced by the explicit rules. The score, policy, VQE optimizer and number of training episodes are as in DreamQAS, but because the internal representation also changes, this is an alternative design rather than a replacement of a single component.

Predictor-QAS uses the same recurrent ensemble architecture and score, but no construction policy. In each training round, it generates 512 complete legal circuits, half at random and half by changing one gate in one of the 32 best verified circuits, and evaluates by VQE, on every partial circuit, the two candidates with the best pessimistic predicted score and the two with the largest disagreement. At deployment, the frozen predictor scores 512 new random legal circuits, and only the circuit with the best pessimistic predicted score is evaluated by VQE, prefix by prefix; its lowest error is the result of that deployment. We report the mean over 100 deployments per seed. Checkpoints are taken after 15,000 training episodes, except on LiH-6q (30,000).

For the same-ensemble comparison, we take, for each task and seed, the checkpoint after 15,000 episodes of a DreamQAS run trained without selective verification and freeze its ensemble. Greedy selection builds a circuit from the empty circuit by adding, at each step, the legal gate with the lowest pessimistic score $\bar y+\beta\sigma$; beam search keeps the ten partial circuits with the lowest score at each depth. Both continue to the maximum depth or until no legal gate remains, and VQE is used only to evaluate the delivered circuit, prefix by prefix, not to choose gates. Both rules are deterministic and deliver one circuit per seed; a stochastic variant of greedy selection, which picks a random legal gate with probability 0.1, is evaluated over 100 episodes (Supplementary Table~\ref{tab:supp_design_controls}). The policy arm uses the policy from the same checkpoint, evaluated over 100 episodes.

For the horizon comparison, the $H=15$ runs are the runs without selective verification used above, and the $H=1$ and $H=5$ runs are trained separately with otherwise the same settings. With $H=15$, the number of imagined steps retained per update varies with truncation, up to 960. The $H=1$ and $H=5$ runs retain a fixed number per update, set to a reference value from the $H=15$ runs (954 for LiH-4q and 960 for LiH-6q); returns are computed on complete imagined sequences before the retained steps are sampled uniformly down to this number.

\subsection{Diagnostic tests}\label{sec:method_diagnostics}

To test whether scoring against the best energy found so far changes which gate additions are rewarded, we recomputed the rewards of the real transitions in the DreamQAS runs using the exact ground-state energy, $r^{E_0}_{t+1}=\log_{10}(E_t-E_0)-\log_{10}(E_{t+1}-E_0)$, and compared them with the real reward $r_{t+1}=y_t-y_{t+1}$ over the final third of the 15,000-episode budget. Spearman correlations use the transitions that change the energy, with tied values given average ranks.

To test the ranking of next gates, the frozen policy builds 15 partial circuits per seed at early, intermediate and final diagnostic checkpoints (Supplementary Table~\ref{tab:supp_action_utility}); all five tasks use the 15,000-episode runs. Circuits with fewer than three legal next gates are discarded; otherwise all legal next gates are tested if there are at most ten, and ten are sampled at random if there are more. Each tested gate is evaluated by VQE with the training optimizer. For each circuit we compute the Spearman correlation between predicted scores and VQE errors and average it over circuits. If $a^\ast$ is the gate with the best predicted score and $e_a$ is the VQE error after gate $a$, the regret is $(e_{a^\ast}-\min_a e_a)/(\max_a e_a-\min_a e_a)$, set to zero when all tested gates tie. The last measure is the fraction of comparisons in which $a^\ast$ is no worse than a gate drawn at random from the same set, counting ties.

To relate disagreement to prediction error, we use the candidates selected for selective verification because of their predicted score, including those also selected for disagreement, collected throughout training on LiH-4q, LiH-6q and BeH$_2$-8q. For each seed separately, candidates are sorted by a retention rule, and for fractions $c=0.10,0.15,\ldots,1.00$ we keep the first $\max(1,\mathrm{round}(cn))$ of the $n$ candidates and record the mean absolute difference between predicted and VQE-derived scores. The rules are lowest disagreement first, best predicted score first, worst predicted score first and a fixed random order. The area under each curve over $0.10\leq c\leq 1.00$, divided by the interval length, summarizes the curve for each seed~\cite{geifman2017selective}; curves and areas are then averaged over seeds.

For the component removals, three variants of DreamQAS each change one component: selective verification is switched off; the disagreement-based controls are switched off ($\beta=0$ and no stopping threshold), while the ranking check, horizon limit and selective verification remain; or the inverse-density weights are replaced by an earlier fixed weighting of the scores (Supplementary Information). All variants use 15,000 training episodes and five seeds.

\subsection{Noise simulations}\label{sec:method_noise}

The noise experiment uses deterministic simulations of noisy expectation values on LiH-4q. A single-qubit depolarizing channel follows every rotation, a two-qubit depolarizing channel follows every CNOT, and the gates that rotate the measurement basis are also noisy. Readout errors are symmetric bit flips, which scale the expectation value of each Pauli term of weight $w$ by $(1-2p_{\mathrm{ro}})^w$. The single-qubit, two-qubit and readout error rates, chosen to resemble calibration data from two IBM devices, are $(1.5\times10^{-4},\,1.1\times10^{-3},\,5.4\times10^{-3})$ in the low-noise setting and $(7.0\times10^{-4},\,7.8\times10^{-3},\,2.0\times10^{-2})$ in the high-noise setting. Amplitude damping, additional dephasing and finite-shot sampling are not modeled. Noise acts on every VQE evaluation during training and frozen evaluation, and the reported error is the noisy energy of the selected partial circuit minus $E_0$. All five methods are trained for 15,000 episodes with five seeds and evaluated over 100 episodes per seed.

%% file: section/supplementary.tex
\clearpage
\section*{Supplementary Information}
\setcounter{section}{0}
\setcounter{subsection}{0}
\setcounter{table}{0}
\setcounter{figure}{0}
\setcounter{equation}{0}
\renewcommand{\thesection}{S\arabic{section}}
\renewcommand{\thesubsection}{\thesection.\arabic{subsection}}
\renewcommand{\thetable}{S\arabic{table}}
\renewcommand{\thefigure}{S\arabic{figure}}
\renewcommand{\theequation}{S\arabic{equation}}
\renewcommand{\theHsection}{S\arabic{section}}
\renewcommand{\theHsubsection}{S\arabic{section}.\arabic{subsection}}
\renewcommand{\theHtable}{S\arabic{table}}
\renewcommand{\theHfigure}{S\arabic{figure}}
\renewcommand{\theHequation}{S\arabic{equation}}

This Supplementary Information provides the task metadata, resolved configurations and extended analyses referred to in the main text. Unless stated otherwise, errors are in mHa, lower values are better, and reported uncertainties are sample standard deviations over five independent training seeds. The exact ground-state energy is used after training to compute the reported error and, as described in Methods, to set the reference energy of the training curriculum.

\section{Molecular systems and search spaces}

The BeH$_2$ task specifications are taken from HamQASBench~\cite{niu2026hamqasbench}; the LiH and H$_2$O molecular setups follow CRLQAS and HyRLQAS~\cite{patel2024curriculum,niu2025hybrid}.

\begin{table}[!ht]
\centering
\footnotesize
\setlength{\tabcolsep}{3pt}
\renewcommand{\arraystretch}{1.1}
\begin{tabular*}{\linewidth}{@{\extracolsep{\fill}}l>{\raggedright\arraybackslash}p{0.33\linewidth}lccr@{}}
\toprule
\textbf{Task} & \textbf{Geometry (\AA)} & \textbf{Basis} & \textbf{Active space} & \textbf{Mapping} & \textbf{$E_0$ (Ha)}\\
\midrule
LiH-4q & Li$(0,0,0)$; H$(0,0,3.4)$ & STO-3G & $(2e,2o)$ & parity & $-7.7890888980$\\
BeH$_2$-6q & Be$(0,0,0)$; H$(0,0,\pm1.33)$ & STO-3G & $(2e,3o)$ & JW & $-14.8615891833$\\
LiH-6q & Li$(0,0,0)$; H$(0,0,2.2)$ & STO-3G & $(2e,3o)$ & JW & $-7.8448790930$\\
BeH$_2$-8q & Be$(0,0,0)$; H$(0,0,\pm1.326)$ & 6-31G & $(2e,4o)$ & JW & $-15.7615037667$\\
BeH$_2$-10q & Be$(0,0,0)$; H$(0,0,\pm1.326)$ & 6-311G & $(2e,5o)$ & JW & $-15.7651241753$\\
\midrule
BeH$_2$-12q & Be$(0,0,0)$; H$(0,0,\pm1.326)$ & cc-pVDZ & $(2e,6o)$ & JW & $-15.7693426958$\\
H$_2$O-8q & O$(0,0,0)$; H$(0,\pm0.757,0.586)$ & STO-3G & $(4e,4o)$ & JW & $-74.9703628391$\\
\bottomrule
\end{tabular*}
\caption{Molecular systems. The five primary tasks are followed by the two extension tasks. JW, Jordan--Wigner.}
\label{tab:supp_molecules}
\end{table}

\begin{table}[!ht]
\centering
\small
\renewcommand{\arraystretch}{1.10}
\begin{tabular*}{\linewidth}{@{\extracolsep{\fill}}lccccl@{}}
\toprule
\textbf{Task} & \textbf{Qubits} & \textbf{Actions} & \textbf{Depth}
& \textbf{Episodes} & \textbf{VQE optimizer}\\
\midrule
LiH-4q & 4 & 24 & 40 & 15,000 & COBYLA ($\leq1{,}000$)\\
BeH$_2$-6q & 6 & 48 & 50 & 15,000 & COBYLA ($\leq1{,}000$)\\
LiH-6q & 6 & 48 & 50 & 15,000 & COBYLA ($\leq1{,}000$)\\
BeH$_2$-8q & 8 & 80 & 50 & 15,000 & ROTOSOLVE (3 sweeps)\\
BeH$_2$-10q & 10 & 120 & 50 & 15,000 & ROTOSOLVE (3 sweeps)\\
BeH$_2$-12q & 12 & 168 & 50 & 15,000 & ROTOSOLVE (3 sweeps)\\
H$_2$O-8q & 8 & 80 & 50 & 15,000 & ROTOSOLVE (3 sweeps)\\
\bottomrule
\end{tabular*}
\caption{Search-space and optimization settings. All experiments use five training seeds. BeH$_2$-12q contains 326 Pauli terms and H$_2$O-8q contains 104.}
\label{tab:supp_search_spaces}
\end{table}

The action dictionary contains three single-qubit rotation gates per qubit and one directed CNOT for every ordered qubit pair, giving $|\mathcal A|=3n_q+n_q(n_q-1)$. Training terminates structurally at the listed depth or through the task curriculum; frozen evaluation always runs to the structural endpoint. The DreamQAS curriculum uses a fixed reference energy set $2.28$~Ha below $E_0$; after the first curriculum update, the stopping threshold depends only on the best energy observed so far (Methods and Supplementary Table~\ref{tab:supp_hyperparameters}).

\section{Implementation and comparison settings}

\begin{table}[!ht]
\centering
\small
\setlength{\tabcolsep}{4pt}
\renewcommand{\arraystretch}{1.08}
\begin{tabular}{@{}lp{0.66\linewidth}@{}}
\toprule
\textbf{Component} & \textbf{Setting}\\
\midrule
Policy
& 3-layer width-512 MLP; Adam, $3\times10^{-5}$; entropy $10^{-3}$;
$\gamma=0.99$; gradient-norm cap 10.\\
Feedback ensemble
& 3 members; 64-dimensional gate embedding; 2-layer GRU, hidden width
256--512; 128-unit SiLU head; randomized-prior coefficient 3.0.\\
Fitting and replay
& Adam, $10^{-3}$; weight decay $10^{-5}$; batch 64; 200 updates per 20
iterations; Huber threshold 1; bootstrap probability 0.8. Replay mix
$0.25/0.35/0.20/0.20$; elite set 1,000; 10 target strata; 40 density bins,
smoothing width 2 and weight cap 50.\\
Imagination and reliability
& 64 recorded episodes from the 200 most recent; horizon 15; $\gamma=0.99$;
$\lambda=0.95$; imagined-loss weight 1.0; warm-up 500 episodes; ranking threshold
0.70; pessimism 1.0; disagreement threshold 0.60.\\
Selective verification
& Every 20 iterations: 5 predicted-best and 5 highest-disagreement circuits.\\
Score and curriculum
& Score scale $m=0.1\mha$. Curriculum reference $2.28$~Ha below $E_0$; initial threshold $2.45$~Ha;
every 500 episodes reset to $0.5\mha$ above the best observed energy; tightened by $0.05\mha$
after every 25 successes, at most ten times.\\
\bottomrule
\end{tabular}
\caption{DreamQAS settings. Further implementation details are available in the public code.}
\label{tab:supp_hyperparameters}
\end{table}

Immediately before each ensemble refresh, the ranking check uses the current, pre-update weights to score all partial circuits in at most the 200 most recent replay episodes. Two index arrays are drawn independently with replacement to form 50,000 pairs; pairs with equal reference energies are excluded. The random generator is reset to seed 0 for each check, so the same index pairs are reused when the number of available partial circuits is unchanged.

One policy update is performed per training iteration. Four real trajectories provide the real REINFORCE term; when the ranking check is passed, the same update also contains the loss from imagined rollouts. One VQE optimization denotes the evaluation of one partial circuit, including the optimization of its parameters. Optimizations in real episodes and in selective verification enter the reported training count, whereas frozen-policy evaluations do not.

\begin{table}[!ht]
\centering
\footnotesize
\setlength{\tabcolsep}{4pt}
\renewcommand{\arraystretch}{1.12}
\begin{tabular*}{\linewidth}{@{\extracolsep{\fill}}>{\raggedright\arraybackslash}p{0.16\linewidth}>{\raggedright\arraybackslash}p{0.62\linewidth}>{\raggedright\arraybackslash}p{0.15\linewidth}@{}}
\toprule
\textbf{Method} & \textbf{Difference from DreamQAS or native protocol} & \textbf{Role}\\
\midrule
DreamQAS-RL & Removes ensemble, replay and imagination. & Matched model-free\\
NoImag & Removes imagined updates and hence selective verification. & Primary control\\
CRLQAS, HyRLQAS & Original rewards and curricula (anchored at $E_0$); same episodes and VQE counting. & Matched RL-QAS\\
\midrule
GQE & Online generative transformer. 1,000 training epochs on LiH-4q, BeH$_2$-6q and LiH-6q, and 1,500 on the larger tasks; warm-up 300 or 450 epochs; 250 sequences per refresh every 25 epochs, half from replay once available. Reported values use the last periodic evaluation (50 generated sequences), at epoch 1,000 or 1,400. COBYLA with 1,000 iterations or ROTOSOLVE with one sweep. & Final quality only\\
TFQAS & One training-free search per seed: 11,250 random candidates are filtered to 1,125 by path count and to 100 by expressibility (2,000 parameter samples, 75 bins)~\cite{sim2019expressibility}; the 100 remaining candidates are evaluated by VQE. COBYLA with 100 iterations or ROTOSOLVE with two sweeps. & Final quality only\\
QuantumDARTS & Differentiable supernet search with 20, 17, 17, 12 and 10 slots per qubit on the five primary tasks (8 on BeH$_2$-12q and 12 on H$_2$O-8q); 500 epochs on the four- and six-qubit tasks, 700 on BeH$_2$-8q and H$_2$O-8q, 800 on BeH$_2$-10q and 900 on BeH$_2$-12q, each with 10 inner steps. Each search delivers one architecture, with at most 40 or 50 gates. & Final quality only\\
\bottomrule
\end{tabular*}
\caption{Compared methods. The first three share the episode budget and VQE counting of DreamQAS. The last three use native protocols with seeds 0--4 and the same Hamiltonians; their budgets are not expressed in VQE optimizations, so they are compared on final circuit quality only.}
\label{tab:supp_methods}
\end{table}

All runs used AWS EC2 g6.24xlarge instances (AMD EPYC 7R13, 48 cores and 96 vCPUs, 363 GiB RAM) with four NVIDIA L4 GPUs (23,034 MiB, compute capability 8.9), one GPU per run and no cross-GPU communication, under Ubuntu 24.04.4 LTS, NVIDIA driver 580.126.16 and Python 3.10.20. Training used PyTorch 2.10.0+cu128 (CUDA 12.8, cuDNN 9.10.2), NumPy 2.2.6, SciPy 1.15.2 and Qulacs 0.6.13 with GPU support; OpenFermion 1.7.1 and PySCF 2.13.1 were used only offline to prepare the molecular Hamiltonians.

\section{Extended policy-quality evaluation}

The two extension tasks test the frozen policies without expanding every mechanism experiment. Their evaluation follows the primary protocol: 100 new episodes per seed, the lowest VQE-evaluated prefix in each episode, and then the mean over five seed-level episode means. BeH$_2$-12q increases the action space to 168, whereas H$_2$O-8q changes both the molecular geometry and active space.

Episode-best evaluation measures whether a policy can produce a useful prefix during one rollout. To expose the quality of the completed architecture as a separate endpoint, Supplementary Table~\ref{tab:supp_final_circuit} also reports the final circuit of the same evaluation episodes, without prefix selection.

\begin{table}[!ht]
\centering
\footnotesize
\setlength{\tabcolsep}{3pt}
\renewcommand{\arraystretch}{1.08}
\begin{tabular*}{\linewidth}{@{\extracolsep{\fill}}lccccc@{}}
\toprule
\multicolumn{6}{@{}l}{\textbf{a}\enspace Final-circuit error, five primary tasks}\\
\textbf{Method} & \textbf{LiH-4q} & \textbf{BeH$_2$-6q} & \textbf{LiH-6q} & \textbf{BeH$_2$-8q} & \textbf{BeH$_2$-10q}\\
\midrule
DreamQAS & $\mathbf{0.162\!\pm\!0.031}$ & $\mathbf{0.975\!\pm\!1.6}$ & $\mathbf{15.4\!\pm\!2.1}$ & $45.1\!\pm\!36$ & $\mathbf{23.1\!\pm\!18}$ \\
\hspace*{1em}NoImag & $1.17\!\pm\!0.20$ & $23.5\!\pm\!12$ & $23.8\!\pm\!6.1$ & $\mathbf{44.6\!\pm\!26}$ & $45.3\!\pm\!52$ \\
\hspace*{1em}DreamQAS-RL & $1.74\!\pm\!0.56$ & $28.5\!\pm\!13$ & $65.3\!\pm\!5.8$ & $372\!\pm\!58$ & $368\!\pm\!35$ \\
CRLQAS & $11.5\!\pm\!2.5$ & $105\!\pm\!13$ & $52.1\!\pm\!6.0$ & $171\!\pm\!80$ & $315\!\pm\!130$ \\
HyRLQAS & $59.6\!\pm\!64$ & $62.5\!\pm\!33$ & $93.5\!\pm\!15$ & $478\!\pm\!180$ & $431\!\pm\!200$ \\
\midrule
\multicolumn{6}{@{}l}{\textbf{b}\enspace Extension tasks}\\
& \multicolumn{2}{c}{\textbf{BeH$_2$-12q}} & \multicolumn{2}{c}{\textbf{H$_2$O-8q}} & \\
\cmidrule(lr){2-3}\cmidrule(lr){4-5}
\textbf{Method} & \textbf{Episode best} & \textbf{Final circuit} & \textbf{Episode best} & \textbf{Final circuit} & \\
\midrule
DreamQAS & $\mathbf{2.00\!\pm\!0.0006}$ & $\mathbf{109\!\pm\!190}$ & $9.89\!\pm\!2.3$ & $\mathbf{28.1\!\pm\!11}$ & \\
\hspace*{1em}NoImag & $3.75\!\pm\!2.6$ & $135\!\pm\!119$ & $20.3\!\pm\!9.0$ & $315\!\pm\!81$ & \\
\hspace*{1em}DreamQAS-RL & $3.69\!\pm\!1.7$ & $631\!\pm\!68$ & $63.3\!\pm\!28$ & $1203\!\pm\!193$ & \\
CRLQAS & $3.50\!\pm\!1.9$ & $156\!\pm\!29$ & $20.3\!\pm\!9.6$ & $812\!\pm\!148$ & \\
HyRLQAS & $2.01\!\pm\!0.021$ & $271\!\pm\!190$ & $\mathbf{7.42\!\pm\!0.00}$ & $231\!\pm\!227$ & \\
GQE & $498\!\pm\!360$ & -- & $2082\!\pm\!1195$ & -- & \\
TFQAS & $222\!\pm\!18$ & -- & $239\!\pm\!14$ & -- & \\
QuantumDARTS & $44.1\!\pm\!38$ & -- & $242\!\pm\!214$ & -- & \\
\bottomrule
\end{tabular*}
\caption{Additional quality evaluations. \textbf{a}, Error of the final circuit of each evaluation episode, from the same frozen-policy episodes as Table~\ref{tab:main_quality}. \textbf{b}, Extension tasks; GQE, TFQAS and QuantumDARTS deliver one circuit, evaluated as in Table~\ref{tab:main_quality}. Indented rows are the internal controls. Bold marks the lowest mean in each column.}
\label{tab:supp_final_circuit}
\end{table}

The final-circuit reduction is deliberately stricter than the primary episode-best metric. DreamQAS has a lower final-circuit mean than CRLQAS, HyRLQAS and DreamQAS-RL on all seven tasks, and than NoImag on all but BeH$_2$-8q, where the two are close ($45.1$ and $44.6\mha$). The larger errors on the extension tasks also show that a policy can visit a useful prefix before its completed circuit deteriorates. For this reason, final-circuit and episode-best results are reported separately.

\section{Complete VQE-efficiency results}

\begin{figure}[!ht]
\centering
\includegraphics[width=\linewidth]{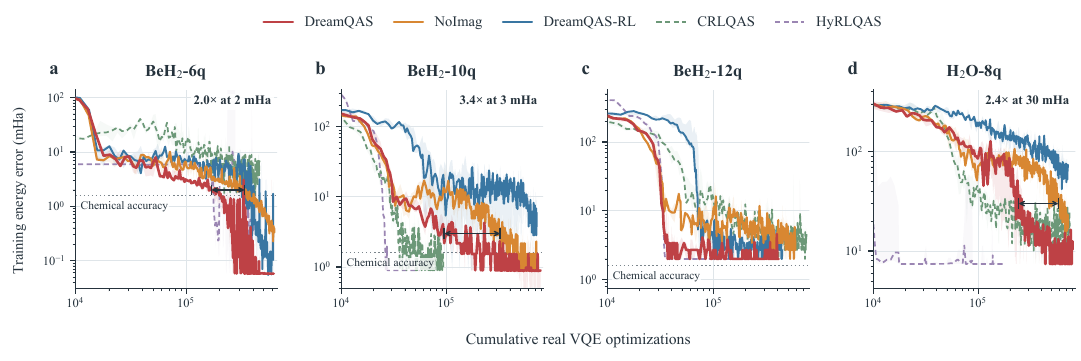}
\caption{Additional training-quality curves against the cumulative number of VQE optimizations. The first two panels complete the five-task primary benchmark; the last two panels show the extension tasks. Curves are seed medians with interquartile ranges of the trailing-100-episode mean. Arrow labels are matched-error gaps between DreamQAS and NoImag at the indicated error levels.}
\label{fig:supp_speed}
\end{figure}

Figure~\ref{fig:supp_speed} extends the main efficiency figure with the remaining two primary tasks and the two extension tasks. Curves use the same trailing-100-episode statistic and include the VQE optimizations used for selective verification. The extension plots are supporting evidence only; the main controlled efficiency statements are based on the five primary tasks.

The matched-error gap at level $q$ is
\begin{equation}
G(q)=\frac{C_{\mathrm{NoImag}}(q)}{C_{\mathrm{DreamQAS}}(q)},
\end{equation}
where $C_m(q)$ is the first cumulative VQE count for which the median curve of method $m$ is at or below $q$ for three consecutive reported points. Thus $G(q)>1$ means that NoImag requires more VQE optimizations to reach the same error. Supplementary Table~\ref{tab:supp_gap_noimag} reports the complete primary-task ladders underlying the curves, for NoImag and for DreamQAS-RL. These ratios describe horizontal gaps between the median curves; they are not averaged across error levels.

\begin{table}[!ht]
\centering
\small
\renewcommand{\arraystretch}{1.02}
\begin{tabular*}{0.86\linewidth}{@{\extracolsep{\fill}}llrrr@{}}
\toprule
\textbf{Task} & \textbf{Target} & \textbf{DreamQAS} & \textbf{NoImag} & \textbf{DreamQAS-RL}\\
\midrule
LiH-4q & 10.0 & 14,650 & 12,985 (0.9) & 6,904 (0.5)\\
& 5.0 & 25,703 & 35,048 (1.4) & 56,302 (2.2)\\
& 2.0 & 90,154 & 124,880 (1.4) & 190,265 (2.1)\\
& 1.0 & 174,152 & 265,387 (1.5) & 296,869 (1.7)\\
& 0.5 & 259,420 & 391,876 (1.5) & 459,904 (1.8)\\
\midrule
BeH$_2$-6q & 10.0 & 17,882 & 15,568 (0.9) & 29,403 (1.6)\\
& 5.0 & 57,777 & 127,602 (2.2) & 134,990 (2.3)\\
& 2.0 & 168,563 & 335,409 (2.0) & 367,075 (2.2)\\
& 1.0 & 235,279 & 448,664 (1.9) & 388,026 (1.6)\\
& 0.5 & 237,665 & 558,238 (2.3) & 388,026 (1.6)\\
\midrule
LiH-6q & 50.0 & 9,645 & 10,458 (1.1) & 8,885 (0.9)\\
& 30.0 & 14,867 & 10,458 (0.7) & 8,885 (0.6)\\
& 20.0 & 210,376 & 420,850 (2.0) & --\\
& 15.0 & 240,615 & 495,850 (2.1) & --\\
& 12.0 & 347,133 & 660,850 (1.9) & --\\
\midrule
BeH$_2$-8q & 20.0 & 24,452 & 75,250 (3.1) & 180,684 (7.4)\\
& 10.0 & 55,249 & 254,175 (4.6) & 419,695 (7.6)\\
& 5.0 & 68,803 & 353,493 (5.1) & --\\
& 3.0 & 193,571 & 437,905 (2.3) & --\\
& 2.5 & 239,639 & 514,904 (2.1) & --\\
\midrule
BeH$_2$-10q & 30.0 & 27,781 & 26,547 (1.0) & 59,798 (2.2)\\
& 15.0 & 30,798 & 31,166 (1.0) & 82,831 (2.7)\\
& 10.0 & 35,846 & 32,190 (0.9) & 97,831 (2.7)\\
& 5.0 & 59,654 & 253,091 (4.2) & 629,743 (10.6)\\
& 3.0 & 94,681 & 322,995 (3.4) & --\\
\bottomrule
\end{tabular*}
\caption{VQE optimizations needed to reach matched error targets (mHa). Parentheses give the ratio to DreamQAS; a dash denotes an unreached target.}
\label{tab:supp_gap_noimag}
\end{table}

The gap depends on the error target and is not necessarily monotonic as the target tightens. Ratios should therefore be read together with their stated error levels, rather than as task-wide constants. A dash indicates that the median curve does not reach the target.

\subsection{Training time by phase}

\begin{figure}[!ht]
\centering
\includegraphics[width=\linewidth]{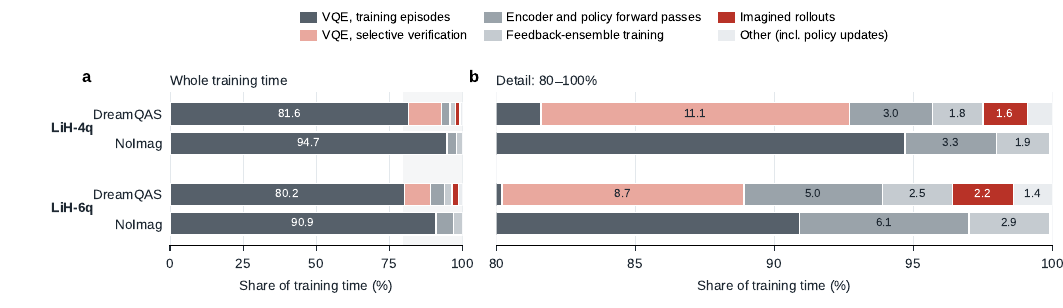}
\caption{\textbf{Share of training time by phase.} Timed runs on LiH-4q and LiH-6q with three seeds per method, over the whole training run. Only within-run shares are shown, because these runs shared a computing node with other jobs and their absolute times are not comparable. Time for selective-verification VQE is estimated from the fraction of VQE optimizations reconstructed from verified circuit lengths ($12.0\%$ and $9.8\%$). \textbf{b}, Enlargement of the last $20\%$ of each bar. The encoder and the feedback ensemble are trained in both DreamQAS and NoImag. All timings are from state-vector simulations and do not measure time on quantum hardware.}
\label{fig:supp_timing}
\end{figure}

To quantify how training time is distributed between VQE and classical computations, we timed separate training runs of DreamQAS and NoImag on LiH-4q and LiH-6q (Supplementary Fig.~\ref{fig:supp_timing}). VQE optimizations dominate the training time of both methods. In DreamQAS, training the feedback ensemble and generating imagined rollouts together take $3$--$5\%$ of the time, and the VQE optimizations used for selective verification are the largest added cost; these optimizations are included in all VQE counts reported in this work.

\clearpage
\section{Design controls}

This section gives the values behind Fig.~\ref{fig:design_controls}, including tasks not shown there. The controls ask whether deterministic circuit transitions should be learned, whether an online circuit predictor can replace policy learning, whether the same ensemble is more useful for training a policy than for selecting gates directly, and whether imagined sequences need to be longer than one step. Methods describes the learned-transition alternative, a DreamerV3-style recurrent state-space model (RSSM) with its own prediction heads, and Predictor-QAS, which scores complete circuits instead of training a policy.

\begin{table}[!ht]
\centering
\footnotesize
\setlength{\tabcolsep}{3pt}
\renewcommand{\arraystretch}{1.08}
\begin{tabular*}{\linewidth}{@{\extracolsep{\fill}}lcccc@{}}
\toprule
\textbf{Variant} & \textbf{LiH-4q} & \textbf{BeH$_2$-6q} & \textbf{LiH-6q} & \textbf{BeH$_2$-8q}\\
\midrule
\multicolumn{5}{@{}l}{\textbf{a}\enspace Circuit transitions}\\
Exact (DreamQAS) & $0.053\!\pm\!0.024$ & $0.058\!\pm\!0.000$ & $11.448\!\pm\!0.471$ & --\\
Learned (RSSM) & $0.411\!\pm\!0.335$ & $1.264\!\pm\!1.653$ & $18.675\!\pm\!6.288$ & --\\
\midrule
\multicolumn{5}{@{}l}{\textbf{b}\enspace Policy training versus Predictor-QAS}\\
Policy (DreamQAS) & $0.053\!\pm\!0.024$ & -- & $11.448\!\pm\!0.471$ & $0.263\!\pm\!0.017^{*}$\\
Predictor-QAS, mean & $9.456\!\pm\!1.603$ & -- & $32.807\!\pm\!1.599^{\dagger}$ & $63.432\!\pm\!16.006$\\
Predictor-QAS, median & $6.762$ & -- & $32.637^{\dagger}$ & $2.175$\\
Predictor-QAS, quartiles & $[4.473, 7.785]$ & -- & $[12.329, 36.885]^{\dagger}$ & $[2.175, 91.765]$\\
\midrule
\multicolumn{5}{@{}l}{\textbf{c}\enspace One frozen ensemble, three uses$^{*}$}\\
Trained policy & $0.073\!\pm\!0.036$ & -- & $11.365\!\pm\!0.647$ & $0.263\!\pm\!0.017$\\
Greedy & $4.280\!\pm\!1.966$ & -- & $36.885\!\pm\!0.000$ & $1.398\!\pm\!1.063$\\
Stochastic greedy & $3.378\!\pm\!0.622$ & -- & $36.733\!\pm\!1.729$ & $1.200\!\pm\!0.274$\\
Beam ($B=10$) & $4.434\!\pm\!0.000$ & -- & $36.941\!\pm\!0.126$ & $1.398\!\pm\!1.063$\\
\midrule
\multicolumn{5}{@{}l}{\textbf{d}\enspace Imagination horizon$^{*}$}\\
$H=1$ & $1.223\!\pm\!1.804$ & -- & $22.226\!\pm\!13.310$ & $1.039\!\pm\!1.025$\\
$H=5$ & $0.137\!\pm\!0.031$ & -- & $11.756\!\pm\!1.012$ & $0.361\!\pm\!0.267$\\
$H=15$ & $0.073\!\pm\!0.036$ & -- & $11.369\!\pm\!0.699$ & $0.263\!\pm\!0.017$\\
\bottomrule
\end{tabular*}
\caption{Design-control results (mHa). Mean $\pm$ SD over five seeds; the Predictor-QAS median and quartiles (25th, 75th percentiles) describe all 500 delivered circuits per task. $^{*}$Runs trained without selective verification. $^{\dagger}$Trained for 30,000 episodes. A dash denotes a setting that was not run.}
\label{tab:supp_design_controls}
\end{table}

The RSSM control is worse for all five paired seeds on each non-saturated LiH task. BeH$_2$-6q is already close to its ansatz floor and does not provide a discriminating transition comparison. On LiH-6q, 87 of 500 RSSM evaluation episodes enter a recurring $36.885\mha$ plateau, compared with none for DreamQAS.

Predictor-QAS has a higher mean deployment error than the corresponding policy control on all three probed tasks, by more than two orders of magnitude on BeH$_2$-8q. Its BeH$_2$-8q distribution is bimodal: 278 of the 500 deployments end at the same $2.175\mha$ circuit, while a substantial fraction return much poorer circuits. Reporting the median alone would therefore obscure its deployment risk.

Predictor-QAS changes both the learned model and the way predictions are used. The same-ensemble control holds the feedback ensemble fixed and changes only how it is used. All entries in each column of block~c use the same checkpoint, a DreamQAS run trained with imagined updates but without selective verification. Greedy selection, stochastic greedy selection (a random legal gate with probability 0.1, otherwise the greedy choice) and beam search act directly on the pessimistic ensemble scores, whereas the trained policy is the policy from the same checkpoint. On both LiH tasks, direct greedy and beam deployment remain far above the learned policy even though the ensemble weights are identical. On BeH$_2$-8q, greedy selection reaches the lowest circuit level, about $0.23\mha$, in two of five seeds and otherwise stops at $2.175\mha$; beam search returns the same circuit as greedy selection for every seed, and stochastic greedy selection averages $1.20\mha$. The trained policy, by contrast, ends below $0.28\mha$ in all five seeds.

The $H=1$ and $H=5$ runs retain a fixed number of imagined steps per update, set to a reference value from the $H=15$ runs, which retain a variable number of up to 960 steps (Methods). Returns are computed on complete imagined sequences before the retained steps are downsampled. Both multi-step settings improve on $H=1$ on all three tasks, and the $H=5$ and $H=15$ means differ by only $0.064\mha$ on LiH-4q, $0.387\mha$ on LiH-6q and $0.098\mha$ on BeH$_2$-8q. These results support learning across several consecutive gate additions but do not identify an optimal horizon. On BeH$_2$-8q, two of the five $H=1$ runs end near $2.15$--$2.18\mha$, whereas all $H=15$ runs end below $0.28\mha$.

\section{Learning targets and reliability analyses}

\subsection{Reward audit against the exact ground-state energy}

DreamQAS learns the frontier-relative score $S(E)$ of Eq.~\eqref{eq:res_score}, that is, the score relative to the best energy observed so far, and does not use the exact ground-state energy $E_0$ in its learning targets or rewards. To check whether this choice changes the direction of the learning feedback, we recomputed two rewards on the same real transitions of the DreamQAS runs:
\begin{equation}
r_{\mathrm{frontier}}(t)=S(E_t)-S(E_{t+1}),\qquad
r_{E_0}(t)=\log_{10}(E_t-E_0)-\log_{10}(E_{t+1}-E_0).
\end{equation}
The second quantity is an offline diagnostic only. The comparison uses the final third of the 15,000-episode budget; transitions whose logged energy does not change are excluded from the rank calculation, and ties receive average ranks.

\begin{table}[!ht]
\centering
\footnotesize
\setlength{\tabcolsep}{3pt}
\renewcommand{\arraystretch}{1.08}
\begin{tabular*}{\linewidth}{@{\extracolsep{\fill}}lcccccc@{}}
\toprule
& \multicolumn{3}{c}{\textbf{Reward audit}} & \multicolumn{3}{c@{}}{\textbf{Independent campaigns (mHa)}}\\
\cmidrule(lr){2-4}\cmidrule(l){5-7}
\textbf{Task} & \textbf{Spearman $\rho$} & \textbf{Sign agreement} & \textbf{Flat} & \textbf{Frontier target} & \textbf{$E_0$ target} & \textbf{Difference}\\
\midrule
LiH-4q & $0.977\!\pm\!0.006$ & $0.9999$ & $0.067$ & $0.053\!\pm\!0.024$ & $0.038\!\pm\!0.057$ & $+0.016$\\
BeH$_2$-6q & $0.973\!\pm\!0.018$ & $1.0000$ & $0.651$ & $0.058\!\pm\!0.000$ & $0.058\!\pm\!0.000$ & $0.000$\\
LiH-6q & $1.000\!\pm\!0.000$ & $0.9996$ & $0.628$ & $13.4\!\pm\!2.4$ & $10.1\!\pm\!2.0$ & $+3.3$\\
BeH$_2$-8q & $0.994\!\pm\!0.013$ & $0.9997$ & $0.902$ & -- & -- & -- \\
BeH$_2$-10q & $0.996\!\pm\!0.007$ & $0.9998$ & $0.903$ & -- & -- & -- \\
\bottomrule
\end{tabular*}
\caption{Training-target checks. Left, frontier-relative and $E_0$-referenced rewards on the same real transitions over the final third of training: late-stage Spearman correlation (mean $\pm$ SD over seeds), sign agreement over energy-changing transitions and fraction of flat transitions. Right, frozen-policy error of independent campaigns trained with each target; the difference is frontier minus $E_0$. Independent campaigns are compared on the three smaller tasks; a dash marks tasks without this comparison. The LiH-6q campaigns use 30,000-episode checkpoints.}
\label{tab:supp_reward_audit}
\end{table}

The two rewards have late-stage rank correlations of $0.973$--$1.000$ and nearly identical signs whenever the energy changes. The agreement in sign follows from the monotonicity of both energy-to-score transformations; the high rank correlation between reward differences is an empirical finding, not guaranteed by monotonicity. Their magnitudes can differ, especially near a moving frontier, so this audit supports preservation of the decision direction rather than numerical equivalence of the rewards.

As a separate sensitivity check, the right half of Supplementary Table~\ref{tab:supp_reward_audit} compares independently trained DreamQAS campaigns on LiH-4q, BeH$_2$-6q and LiH-6q using the frontier-relative target and a target computed from exact $E_0$. Because the stochastic trajectories differ between campaigns, these rows are descriptive and are not paired target ablations.

The frontier-relative campaign is similar on LiH-4q and BeH$_2$-6q and worse on LiH-6q. Thus removing $E_0$ does not impose a uniform quality penalty, but neither does this cross-campaign comparison establish exact equivalence.

\subsection{Ranking of next gates}

Methods describes the ranking test. Supplementary Table~\ref{tab:supp_action_utility} reports all five tasks. A regret of zero can arise either because the predicted gate is best or because all tested gates tie, which is common on BeH$_2$-8q and BeH$_2$-10q; the main text therefore shows the final gate-selection measures only for the other three tasks. The Spearman statistic uses ordinal ranks and retains exact ties.

\begin{table}[!ht]
\centering
\small
\setlength{\tabcolsep}{2.5pt}
\renewcommand{\arraystretch}{1.12}
\begin{tabular*}{\linewidth}{@{\extracolsep{\fill}}lccccc@{}}
\toprule
\multirow{2}{*}{\textbf{Task}} & \multicolumn{3}{c}{\textbf{Rank correlation with VQE}} & \multicolumn{2}{c@{}}{\textbf{Final gate selection}}\\
\cmidrule(lr){2-4}\cmidrule(l){5-6}
& \shortstack{\textbf{After}\\\textbf{warm-up}} & \textbf{25\% budget} & \textbf{Final} & \textbf{Regret} $\downarrow$ & \shortstack{\textbf{No worse}\\\textbf{than random} $\uparrow$}\\
\midrule
LiH-4q & $0.234\!\pm\!0.106$ & $0.469\!\pm\!0.034$ & $0.683\!\pm\!0.092$ & $0.026\!\pm\!0.046$ & $0.942\!\pm\!0.040$\\
BeH$_2$-6q & $0.078\!\pm\!0.072$ & $0.175\!\pm\!0.089$ & $0.226\!\pm\!0.073$ & $0.152\!\pm\!0.056$ & $0.649\!\pm\!0.106$\\
LiH-6q & $0.155\!\pm\!0.055$ & $0.330\!\pm\!0.036$ & $0.436\!\pm\!0.087$ & $0.054\!\pm\!0.048$ & $0.707\!\pm\!0.078$\\
BeH$_2$-8q & $0.074\!\pm\!0.089$ & $0.417\!\pm\!0.115$ & $0.515\!\pm\!0.099$ & $(0.000\!\pm\!0.000)^{\dagger}$ & $0.973\!\pm\!0.060$ \\
BeH$_2$-10q & $0.000\!\pm\!0.051$ & $0.384\!\pm\!0.059$ & $0.411\!\pm\!0.054$ & $(0.000\!\pm\!0.000)^{\dagger}$ & $0.978\!\pm\!0.031$ \\
\bottomrule
\end{tabular*}
\caption{Action-ranking probe for all five tasks. Mean $\pm$ sample SD over five seeds. The three checkpoints are at episodes 532, 3,748 and 15,000 on every task, using the nearest available checkpoint for the first two. Wins include ties, so a random choice scores at least 0.5 in the last column. $^{\dagger}$Zero regret does not distinguish tied best actions; on BeH$_2$-8q and BeH$_2$-10q many tested gates reach the same VQE energy.}
\label{tab:supp_action_utility}
\end{table}

\subsection{Risk--coverage and component ablations}

The risk--coverage experiment (Methods) uses candidates selected for predicted value, not for disagreement, so the sample is not constructed by the uncertainty measure being tested. It contains 935, 1,870 and 935 candidates per seed on LiH-4q, LiH-6q and BeH$_2$-8q, collected throughout training. Coverage is the fraction retained after rejecting the largest disagreement values, and risk is the mean absolute error in the frontier-relative target among retained candidates. The area under this curve (AURC) is computed per seed and then averaged; lower values indicate that a rejection rule removes unreliable predictions earlier.

\begin{table}[!ht]
\centering
\footnotesize
\setlength{\tabcolsep}{3.2pt}
\renewcommand{\arraystretch}{1.12}
\begin{tabular*}{\linewidth}{@{\extracolsep{\fill}}lcccccc@{}}
\toprule
\textbf{Task} & \textbf{Full risk} & \shortstack{\textbf{Risk at}\\\textbf{50\%}}
& \textbf{Reduction} & \shortstack{\textbf{AURC}\\\textbf{disagreement}}
& \shortstack{\textbf{AURC}\\\textbf{predicted best}}
& \shortstack{\textbf{AURC}\\\textbf{random}}\\
\midrule
LiH-4q & $0.3617$ & $0.2760$ & $24\%$ & $0.2927$ & $0.3009$ & $0.3686$\\
LiH-6q & $0.3745$ & $0.2759$ & $26\%$ & $0.2900$ & $0.5264$ & $0.3720$\\
BeH$_2$-8q & $0.2357$ & $0.1270$ & $46\%$ & $0.1301$ & $0.2060$ & $0.2332$ \\
\bottomrule
\end{tabular*}
\caption{Risk--coverage summary. Risk is measured in units of the score $y$ (Eq.~\eqref{eq:res_score}), not mHa. The 50\% column retains the half of candidates with the lowest ensemble disagreement; lower AURC is better.}
\label{tab:supp_risk_coverage}
\end{table}

Disagreement rejection lowers risk relative to random rejection on all three tasks. Per seed, disagreement gives a lower AURC than random retention in all 15 task--seed combinations and a lower AURC than predicted-best retention in 13 of 15; the exceptions are LiH-4q seed 0 ($0.313$ versus $0.301$) and BeH$_2$-8q seed 3 ($0.366$ versus $0.274$). It is not uniformly better than ranking by the predicted target itself: the clearest separation occurs on LiH-6q, where retaining the predicted-best half raises risk to $0.5011$, whereas disagreement-based retention lowers it to $0.2759$. At the fixed disagreement threshold of $0.60$, the retained fraction varies by task: approximately 98\%, 99\% and 76\% of value-selected candidates on LiH-4q, LiH-6q and BeH$_2$-8q, respectively. Disagreement also enters continuously through the pessimistic feedback term even when the hard threshold rarely terminates a rollout.

\begin{table}[!ht]
\centering
\small
\setlength{\tabcolsep}{2.5pt}
\renewcommand{\arraystretch}{1.10}
\begin{tabular*}{\linewidth}{@{\extracolsep{\fill}}lccccc@{}}
\toprule
\textbf{Arm} & \textbf{LiH-4q} & \textbf{BeH$_2$-6q}
& \textbf{LiH-6q} & \textbf{BeH$_2$-8q} & \textbf{BeH$_2$-10q}\\
\midrule
DreamQAS & $\mathbf{0.053\!\pm\!0.024}$ & $\mathbf{0.058\!\pm\!0.000}$
& $11.4\!\pm\!0.47$ & $\mathbf{0.256\!\pm\!0.033}$
& $\mathbf{0.528\!\pm\!0.327}$\\
$-$selective verification & $0.073\!\pm\!0.036$ & $0.058\!\pm\!0.000$
& $11.4\!\pm\!0.70$ & $0.263\!\pm\!0.017$ & $0.667\!\pm\!0.550$\\
$-$density weighting & $0.125\!\pm\!0.070$ & $0.058\!\pm\!0.000$
& $\mathbf{11.1\!\pm\!0.63}$ & $0.315\!\pm\!0.121$ & $0.867\!\pm\!0.441$\\
$-$disagreement controls & $0.053\!\pm\!0.048$ & $0.058\!\pm\!0.000$
& $11.2\!\pm\!1.00$ & $0.275\!\pm\!0.026$ & $0.659\!\pm\!0.557$\\
NoImag & $0.221\!\pm\!0.062$ & $0.328\!\pm\!0.120$
& $13.6\!\pm\!3.50$ & $1.929\!\pm\!1.766$ & $1.399\!\pm\!0.613$\\
\bottomrule
\end{tabular*}
\caption{Component removals. Bold marks the lowest displayed mean for each task. Removing the disagreement controls sets the pessimism coefficient to zero and removes the stopping threshold, while the ranking check, horizon limit and selective verification remain. The density-weighting variant replaces the inverse-density weights with an earlier fixed weighting rather than removing weighting. NoImag makes no imagined updates and therefore performs no selective verification.}
\label{tab:supp_ablation}
\end{table}

The density-weighting variant uses $w_{\mathrm{legacy}}(y)=\min\{20,\max[1,5/(10^y+0.3)]\}$ in place of the inverse-density weights; the replay mixture is unchanged. In the full method, weights are computed from 40 target bins, with Gaussian smoothing of width 2 followed by adding one to the smoothed counts. Their inverse square roots are normalized by the mean across bins and capped at 50.

No single component is required for imagination to improve on NoImag: every variant does so on all five tasks. The density-weighting variant is worse on LiH-4q and BeH$_2$-10q and slightly better on LiH-6q, and on BeH$_2$-8q and BeH$_2$-10q every variant has a higher mean error than full DreamQAS, by amounts comparable to the spread across seeds. BeH$_2$-6q reaches the same circuit floor under every variant and does not distinguish them. The comparison with NoImag captures the combined contribution of imagined updates and the verification mechanism that depends on them.

\clearpage
\section{Noise simulations and aggregation}

\subsection{Noise settings}

The noise study uses deterministic simulations of noisy expectation values without finite-shot sampling (Methods). The two settings combine single-qubit and two-qubit depolarizing errors with readout errors, at rates chosen to resemble calibration data from two IBM devices; they are controlled lower- and higher-noise settings rather than models of a particular device. The low-noise setting uses one-qubit, two-qubit and readout error rates of $1.5\times10^{-4}$, $1.1\times10^{-3}$ and $5.4\times10^{-3}$, and the high-noise setting uses $7.0\times10^{-4}$, $7.8\times10^{-3}$ and $2.0\times10^{-2}$.

All five methods use the same LiH-4q Hamiltonian, episode budget and five training seeds. Each frozen checkpoint is evaluated over 100 new episodes using the noisy energy signal available to that method. The reported circuit in each episode is therefore selected without access to a noiseless energy. The reported error is the noisy energy of the selected partial circuit minus $E_0$ (main-text Table~\ref{tab:noise}), so it includes the energy shift caused by noise.

\subsection{Aggregation and censoring}

The independent training seed is the experimental unit throughout. For frozen-policy quality, the 100 episode-level values are first averaged within each seed; tables then report the mean and sample standard deviation across five seed-level means. Predictor-QAS quartiles are the exception explicitly identified in Table~\ref{tab:supp_design_controls}: they describe all delivered circuits to expose its non-Gaussian deployment distribution. Risk--coverage curves and their areas are computed separately for each seed before aggregation.

Efficiency traces use the trailing mean over 100 real episodes and are sampled every 50 episodes. The figure annotations and matched-error ladders use crossings of the median curves, requiring three consecutive samples at or below the target. Separate seed-level crossing summaries in the machine-readable results apply this criterion to individual seeds: unreached seeds are right-censored, and crossing-count medians use only reached seeds, with reach counts reported alongside them. These seed-level summaries are not the source of the figure annotations or ladder ratios.

Resolved configurations, seed-level outputs, analysis scripts and molecular generation files are available at \url{https://github.com/jiayangniu/DreamQAS.git}.